\documentclass[twoside,11pt]{article}

\usepackage{blindtext}

\usepackage{jmlr2e}

\usepackage[utf8]{inputenc} 
\usepackage[T1]{fontenc}    
\usepackage{hyperref}       
\usepackage{url}            
\usepackage{booktabs}       
\usepackage{amsfonts}       
\usepackage{nicefrac}       
\usepackage{microtype}      
\usepackage{xcolor}         

\usepackage{mathtools}
\usepackage{fancyhdr,graphicx,amsmath,amssymb}
\usepackage[ruled,vlined]{algorithm2e}
\usepackage[e]{esvect}

\usepackage{algorithm2e}

\usepackage{varwidth}
\newtheorem{condition}{Condition}

\usepackage{multirow}
\usepackage{comment}
\newtheorem{exmp}{Example}

\newcommand{\pkg}[1]{\textsf{#1}}

\newcommand{\norm}[1]{\left\lVert#1\right\rVert}

\DeclareMathOperator*{\argmax}{arg\,max}

\usepackage{lastpage}

\ShortHeadings{Constructive Approximation and Sure Convergence}{Chi}
\firstpageno{1}

\begin{document}

\title{\bf Constructive Universal Approximation and Sure Convergence for Multi-Layer Neural Networks}
	
	\author{Chien-Ming Chi\thanks{
			Chien-Ming Chi is Assistant Research Fellow, Institute of Statistical Science, Academia Sinica, Taipei 11529, Taiwan (Email: \textit{xbbchi@stat.sinica.edu.tw}). %
			This work was supported by grant 113-2118-M-001-008-MY2 from the National Science and Technology Council, Taiwan.	}
		\hspace{.2cm}\\
		Academia Sinica}
	\date{}
	
\editor{}

\maketitle

\begin{abstract}
We propose o1Neuro, a new neural network model built on sparse indicator activation neurons, with two key statistical properties. 
(1) Constructive universal approximation: At the population level, a deep o1Neuro can approximate any measurable function of $\boldsymbol{X}$, while a shallow o1Neuro suffices for additive models with two-way interaction components, including XOR and univariate terms, assuming $\boldsymbol{X} \in [0,1]^p$ has bounded density. Combined with prior work showing that a single-hidden-layer non-sparse network is a universal approximator, this highlights a trade-off between activation sparsity and network depth in approximation capability. (2) Sure convergence: At the sample level, o1Neuro’s optimization reaches an optimal model with probability approaching one after sufficiently many update rounds, and we provide an example showing that the required number of updates is well bounded under linear data-generating models. 
Empirically, o1Neuro is compared with XGBoost, Random Forests, and TabNet for learning complex regression functions with interactions, demonstrating superior predictive performance on several benchmark datasets from OpenML and the UCI Machine Learning Repository with $n = 10{,}000$, as well as on synthetic datasets with $100 \le n \le 20{,}000$.

\end{abstract}


\section{Introduction}

Predictive models based on neural networks \citep{lecun2015deep} lie at the core of recent artificial intelligence advancements, including AlphaGo \citep{silver2016mastering}, GPT-3 \citep{brown2020language}, and AlphaFold \citep{jumper2021highly}. Their high prediction accuracy has been extensively studied and supported from theoretical perspectives, spanning from approximation results for single-hidden-layer networks \citep{hornik1989multilayer, cybenko1989approximation, barron1993universal} to advanced universal approximation theories for multilayer networks \citep{bauer2019deep, schmidt2020nonparametric, jiao2023deep}.

However, despite their strong empirical performance and universal approximation ability, neural networks can sometimes fall short in practice. Numerous formal studies have shown that neural networks may not outperform other prediction methods on tabular data \citep{grinsztajn2022tree, mcelfresh2023neural, shwartz2022tabular}. In particular, multilayer neural networks have been found to be consistently outperformed by tree-based models \citep{grinsztajn2022tree}, a result that appears to conflict with their universal approximation capabilities. This gap between theory and practice has been a subject of extensive investigation \citep{adcock2021gap, grohs2024proof}. A plausible explanation for this gap may be the failure to achieve complete model optimization, a condition that universal approximation theory assumes \citep{jiao2023deep} but is rarely guaranteed in practice \citep{fan2020selective}.
Whether existing neural networks possess universal approximation capability when training processes are taken into account, often referred to as constructive approximation~\citep{gentile2024approximation}, and how to effectively realize this capability remain significant open problems.

This work introduces o1Neuro, a neural network framework based on sparse indicator activations. Its design is inspired by greedy algorithms~\citep{temlyakov2000weak, devore1996some, jones1992simple, barron1993universal} and boosting methods~\citep{friedman2001greedy}, and the framework illustrates how greedy algorithm and boosting concepts can be extended to optimize multi-layer neural networks. When properly tuned at the population level, o1Neuro can approximate a broad class of data-generating functions, including models of the form 
\(\mathbb{E}(Y \mid \boldsymbol{X}) = h(X_1, \dots, X_k)\) 
for any measurable function \(h: [0,1]^k \to \mathbb{R}\), where the \(p\)-dimensional input \(\boldsymbol{X}\) has a bounded density. In particular,
\begin{itemize}
    \item When the number of hidden layers \(L = 2 + \lceil \log_2 p \rceil\), o1Neuro can approximate \emph{any measurable function} on \([0,1]^p\).
    \item When \(L = 3\), it can approximate any additive model of the form
    $
    \mathbb{E}(Y \mid \boldsymbol{X}) = \sum_{r=1}^{R_0} h_r(X_{2r-1}, X_{2r}),
    $
    which may include multiple XOR-type interactions as well as univariate functions.
\end{itemize}
These results have two important implications.  First, from a theoretical perspective, Theorem~3.8 of \citep{barron2008approximation} similarly shows that a one-hidden-layer \emph{non-sparse} neural network, constructed via an orthogonal (or relaxed) greedy algorithm, can approximate any measurable function of $\boldsymbol{X}$. Taken together, our results and theirs reveal a trade-off between activation sparsity and network depth in achieving strong approximation capability.

Second, from a practical standpoint, Theorem~\ref{lemma1} shows that o1Neuro, even when designed with both sparse activations~\citep{lee1996efficient} to mitigate the curse of dimensionality and shallow networks to enable faster optimization, retains strong approximation power. Specifically, shallow o1Neuro configurations can still approximate a rich class of commonly seen functions, including additive models~\citep{friedman2001greedy} and interaction effects~\citep{bien2013lasso, cox1984interaction}, sufficient for most prediction applications. This ensures that the practical design choices do not compromise approximation capability. In contrast, most existing approximation theories~\citep{schmidt2020nonparametric, bauer2019deep, yarotsky2018optimal, jiao2023deep, barron2008approximation} focus on universal approximation while largely overlooking these practical optimization considerations.

We study the optimization of o1Neuro (in the algorithmic sense), complementing its approximation results. At the sample level, o1Neuro updates each output neuron (last hidden-layer neurons) via a variant of the iterative greedy algorithm, designed for networks with sparse structures, idle neurons, and indicator-based activations (``neurons'' and ``activation functions'' are treated equivalently in this paper). The sample-level optimization of o1Neuro is guaranteed to converge probabilistically to a network that serves as the sample counterpart of the population-level optimal o1Neuro, given a sufficient number of update rounds. This property is referred to as \textit{sure convergence}. Moreover, we show through an example that, under linear data-generating processes, the required number of rounds admits a reasonably tight upper bound.

The proposed o1Neuro is empirically compared with XGBoost~\citep{chen2016xgboost}, Random Forests~\citep{breiman2001random}, and TabNet~\citep{arik2021tabnet} for learning complex regression functions in Sections~\ref{Sec6b}--\ref{Sec7b}. It demonstrates superior predictive performance on several benchmark datasets from OpenML~\citep{OpenML2013} and the UCI Machine Learning Repository~\citep{Dua:2017}, as well as in synthetic experiments. Overall, o1Neuro strikes a balance between theory and practice, appealing to researchers and practitioners interested in constructive deep learning theory, network inference, and accurate mean regression prediction.

The remainder of the paper is organized as follows. Section~\ref{Sec1.1b} reviews related work on neural networks, focusing on two aspects closely related to our study: approximation theory and optimization convergence. Section~\ref{Sec2b} introduces the population- and sample-level o1Neuro, while Section~\ref{Sec3b} presents its constructive universal approximation theory and sure convergence property. Section~\ref{Sec5b} describes training techniques for accelerating hyperparameter optimization and the training schedule used. All technical proofs are provided in the online Supplementary Material.

\subsection{Related Work}\label{Sec1.1b}

To the best of our knowledge, o1Neuro is the only approach that achieves both optimization convergence and universal approximation guarantees under a reasonable setup for multi-layer networks. In contrast, Adam~\citep{kingma2015adam, li2023convergence} and stochastic gradient descent~\citep{fehrman2020convergence} ensure optimization convergence but do not address model asymptotics. Gradient-flow (gradient-descent) methods~\citep{gentile2024approximation} study both optimization and approximation but are limited to single-hidden-layer networks and one-dimensional data-generating functions. The Neural Tangent Kernel (NTK) framework~\citep{jacot2018neural} studies gradient descent, providing guarantees of optimization convergence under strong over-parameterization assumptions (infinitely wide networks). Adaptive annealing~\citep{barron2007adaptive} lacks formal optimization convergence guarantees. Finally, conventional greedy algorithms establish approximation results but either do not specify a concrete optimization procedure~\citep{barron2008approximation} or rely on optimization schemes with impractically high time complexity~\citep{lee1996efficient, farago1993strong}.

We next review approximation and optimization results in separate subsections.

\subsubsection{Universal Approximation in Neural Networks}\label{Sec4.1b}

We begin by reviewing key results in nonconstructive universal approximation theory, keeping the discussion concise due to space constraints. Classical results for single-hidden-layer networks were first established by \citep{cybenko1989approximation, hornik1989multilayer}, showing that such networks can approximate any continuous function on compact domains. These theories primarily describe the expressive power of neural network function classes without addressing how such functions can be computed in practice, a perspective often referred to as algorithm-independent control~\citep{fan2020selective, gentile2024approximation}, or nonconstructive approximation theory. Least-squares estimators~\citep{jiao2023deep, gyorfi2002distribution} are commonly assumed in related work. This line of research, together with the empirical success of deep learning~\citep{krizhevsky2012imagenet}, motivates the study of how modern multilayer architectures can improve approximation convergence~\citep{bauer2019deep, jiao2023deep, schmidt2020nonparametric}.

The universality of single-hidden-layer networks also inspired early training methods, including greedy algorithms~\citep{jones1992simple} and subsequent works~\citep{herrmann2022constructive, siegel2022optimal, barron2008approximation, devore1996some, barron1993universal}. For instance, \citep{barron2008approximation} applies orthogonal and relaxed greedy algorithms to single-hidden-layer networks, sequentially selecting neurons to achieve universal approximation. Their framework focuses on optimizing superpositions of ridge functions without addressing deeper networks, and their optimization is solved approximately using methods such as adaptive annealing~\citep{barron2007adaptive}, without convergence guarantees.

Recently, constructive approximation theories have attracted attention by providing guarantees alongside explicit training procedures. However, existing work remains preliminary, typically limited to shallow networks and simple data-generating functions~\citep{gentile2024approximation, jentzen2022proof}, even when explicitly considering gradient-based updates and network architectures.

\subsubsection{Optimization Convergence in Neural Networks}\label{Sec4.2b}

Training neural networks typically involves minimizing the empirical $L_2$ loss over a given class of network functions. Among the most widely adopted training algorithms are adaptive gradient methods such as Adam and its variants~\citep{kingma2015adam, hinton2012neural, duchi2011adaptive}, which have recently been shown to converge under certain conditions~\citep{zhang2022adam, li2023convergence, defossez2020simple}. Nevertheless, both theoretical~\citep{jentzen2024non} and empirical~\citep{choromanska2015loss} studies indicate that gradient-based optimization often converges to local minima. Although many local minima in sufficiently large or overparameterized networks can still lead to good generalization performance~\citep{choromanska2015loss}, a substantial gap remains between existing theoretical guarantees and the practical behavior of neural networks~\citep{adcock2021gap, grohs2024proof}.

A related line of research analyzes optimization convergence and generalization in infinitely wide neural networks under the NTK framework~\citep{jacot2018neural, cao2019generalization, arora2019exact, xu2024overparametrized}, including extensions to adaptive methods such as Adam~\citep{malladi2023kernel}. In parallel, \citep{mei2018mean, mei2019mean} provide a mean-field perspective, studying the distributional dynamics of parameters updated via stochastic gradient descent in overparameterized single-hidden-layer networks and characterizing conditions under which local or global optima can be attained. While NTK and mean-field theories have significantly advanced the theoretical understanding of neural network optimization, their reliance on assumptions such as infinite or extremely wide architectures limits their applicability to practical settings with moderate network width.

Several greedy algorithms have been proposed for training single-hidden-layer networks, including sparse architectures with indicator activations~\citep{farago1993strong, lee1996efficient}, conceptually similar to a single-hidden-layer o1Neuro. However, optimization is not fully addressed; for instance, CONSTRUCT~\citep{lee1996efficient} systematically explores all hyperplane splits, limiting scalability on large datasets, while adaptive annealing~\citep{barron2007adaptive} uses a heuristic approach without guaranteed convergence.

\subsection{Notation}

Let $(\Omega, \mathcal{F}, \mathbb{P})$ be a probability space. Random vectors are denoted by bold-faced symbols, such as $\boldsymbol{X}$ or $\boldsymbol{X}_j$, while random variables are denoted by $Y$, $Z$, or $X_j$. A constant $k$-dimensional vector is represented by $\vv{x} = (x_1, \dots, x_k)^{\top} \in \mathbb{R}^k$, with its squared $L_2$-norm defined by $\norm{\vv{x}}_2^2 = \sum_{j=1}^k x_j^2$ and its $L_0$-norm by $\norm{\vv{x}}_0 = \texttt{\#} \{j : |x_j| > 0\}$. The indicator function, denoted by $\boldsymbol{1}\{\cdot\}$, equals 1 if its condition is true and 0 otherwise. The smallest integer greater than $x$ is denoted by $\lceil x \rceil$, while the largest integer less than $x$ is denoted by $\lfloor x \rfloor$. We adopt the convention that $\sum_{i=a}^{b} c_i = 0$ when $b < a$, and define any division by zero to be zero.

\section{o1Neuro: Population and Sample Algorithms}\label{Sec2b}

We present the population-level (Section~\ref{Sec2.1b}) and sample-level (Section~\ref{Sec2.2b}) formulations of o1Neuro separately to clearly illustrate our theoretical contributions to both approximation theory and optimization in multi-layer neural networks.

\subsection{Population o1Neuro Model}\label{Sec2.1b}

The o1Neuro network function consists of a neural network with $L$ hidden layers (hereafter referred to as layers), where the $l$th layer has $p_{l}$ activation functions (also called neuron functions, or simply neurons). The name ``01Neuro'' reflects the use of indicator activations; it is written as ``o1Neuro'' since many programming languages disallow identifiers starting with a digit. For $l\in \{1, \dots, L\}$ and $h\in \{1, \dots, p_{l}\}$, define the activation function $f_{l, h}: \mathbb{R}^{p}\mapsto \{0, 1\}$  such that $f_{l, h}(\vv{x})  = \boldsymbol{1}\{ \vv{w}_{l, h}^{\top} \vv{f}_{l-1}(\vv{x}) > c_{l, h} \}$ and
$\vv{f}_{l-1}(\vv{x})  = (f_{l-1, 1}(\vv{x}), \dots, f_{l-1, p_{l-1}}(\vv{x}) )^{\top}$, with 
$f_{0, h}(\vv{x}) = x_{h}$ where $\vv{x}= (x_1, \dots, x_{p })^{\top}\in \mathbb{R}^{p}$, $p_{0} = p$, and the \textit{population  parameter space} defined by
\begin{equation}
    \label{restriction.1}
    \begin{split}
    \vv{w}_{l, h} & \in \mathbb{R}^{p_{l-1}}, \quad \norm{\vv{w}_{l, h}}_2 = 1,\quad\norm{\vv{w}_{l, h}}_0 \le 2, \quad \text{ and }  \quad c_{l, h} \in \mathbb{R}.    
    \end{split}
\end{equation}
Two neurons are \textit{directly connected} if the upper neuron assigns a nonzero weight to the lower; they are \textit{connected} if joined by a path of direct connections. Each output neuron \( f_{L, h} \) induces a \textit{subnetwork} consisting of all neurons connected to it. See Figure~\ref{fig:aro1neuro1} for a graphical illustration of the o1Neuro architecture. 
\begin{remark}
In our experiments, increasing the sparsity  $\norm{\vv{w}_{l,h}}_0 \le w_0$ beyond $w_0 = 2$ does not improve predictive performance but increases computational cost, so we restrict $w_0 = 2$ to simplify notation. Moreover, o1Neuro defines its output layer as the final hidden layer in a conventional neural network architecture.
\end{remark}

\subsubsection{Population-Level Optimization of o1Neuro} \label{Sec2.1.1}
Let $Y$ denote the response variable and $\boldsymbol{X}$ the $p$-dimensional feature vector. Starting from an arbitrarily initialized o1Neuro model, we define the population o1Neuro estimator of $\mathbb{E}[Y \mid \boldsymbol{X}]$, denoted $\widetilde{m}: \mathbb{R}^p \to \mathbb{R}$, as
 \begin{equation}
    \label{population.01}
    \widetilde{m}(\vv{x}) = \gamma\sum_{h=1}^{p_{L}} \sum_{l=0}^{1} \widetilde{a}_{lh}(f_{L, h}) \times \boldsymbol{1}\{f_{L, h} (\vv{x}) = l\}
\end{equation}
for every $\vv{x} \in [0, 1]^{p}$, where $\gamma\in (0, 1]$ is a hyperparameter of boosting learning rate, and we recursively define $\widetilde{R}_{h} = \widetilde{R}_{h-1} - \gamma \sum_{l=0}^{1} \widetilde{a}_{lh}(f_{L, h}) \times \boldsymbol{1}\{f_{L, h} (\boldsymbol{X}) = l\}$ for $h\in \{1, \dots, p_{L}\}$ with $\widetilde{a}_{lh}(f) = \frac{ \mathbb{E}[\widetilde{R}_{h-1} \times \boldsymbol{1}\{f (\boldsymbol{X}) = l\} ] }{\mathbb{P}(f (\boldsymbol{X}) = l) }$, and $\widetilde{R}_{0} = Y$.

 For any $f_{L,h}$'s, \eqref{population.01} is well-defined.  
We focus on the population-level optimal predictor recursively defined as follows. Each $f_{L,h}$ maximizes $\widetilde{W}_h(f_{L,h})$ subject to the parameter constraint \eqref{restriction.1} for $h \in \{1, \dots, p_L\}$, with ties broken randomly.
Here, for any $f: \mathbb{R}^p \to \{0,1\}$, we define
\[
\widetilde{W}_h(f) \coloneqq 
\sum_{l=0}^1 
\frac{\left|\mathbb{E}\!\left[\widetilde{R}_{h-1} \times\boldsymbol{1}\{f(\boldsymbol{X}) = l\}\right]\right|^2}
{\mathbb{P}(f(\boldsymbol{X}) = l)}.
\]
In addition, during the optimization of $f_{L,h}$, the parameters of the subnetworks associated with the preceding output neurons $f_{L,1}, \dots, f_{L,h-1}$ are kept fixed. Although each neuron is updated at most once, it can still serve as input to multiple neurons in higher layers.

\begin{remark}
The above maximization problem is equivalent to minimizing the mean squared loss, given by the left-hand side of the following identity, valid for any $f: \mathbb{R}^p \to \{0, 1\}$.
\begin{equation*}
    \begin{split}
         \mathbb{E}\big\{\big[\widetilde{R}_{h-1} - \gamma\sum_{l=0}^{1} \widetilde{a}_{lh}(f) \times \boldsymbol{1}\{f (\boldsymbol{X}) = l\} \big]^2 \big\}
        & = \mathbb{E}(\widetilde{R}_{h-1}^2) - \gamma(2-\gamma) \widetilde{W}_h(f).        
    \end{split}
\end{equation*}
\end{remark}

\subsection{Sample-Level o1Neuro Model}\label{Sec2.2b}

Let \(\{(\boldsymbol{X}_i, Y_i)\}_{i=1}^n\) denote the training samples. Initialize the o1Neuro model (Section~\ref{Sec2.1b}) with all weights and biases set to zero. The \textit{sample network parameter space} is defined as
\begin{equation}
    \label{restriction.2}
    \begin{split}
    \vv{w}_{l, h} & \in \mathbb{R}^{p_{l-1}}, \quad \norm{\vv{w}_{l, h}}_2 = 1,\quad \text{ and } \quad \norm{\vv{w}_{l, h}}_0 \le 2, \\
    c_{1, h}& \in \{ \vv{w}_{1, h}^{\top}\boldsymbol{X}_i  : 1\le i \le n\}, \quad \text{ and } \quad c_{l, h} \in \{ \vv{w}_{l, h}^{\top} \vv{e} : \vv{e} \in \{0, 1\}^{p_{l-1}}\} \text{ for } l  >1.
    \end{split}
\end{equation}
A random draw of $\vv{w}_{l,h}$ from space~\eqref{restriction.2} yields 
$\|\vv{w}_{l,h}\|_0 = 1$ and $\|\vv{w}_{l,h}\|_0 = 2$ with equal probability, each being $\frac{1}{2}$.

The sample-level o1Neuro model is based on the boosting machine of \citep{friedman2001greedy} and is defined by 
\begin{equation}
    \label{sample.predictor.1}
    \widehat{m}(\vv{x}) = \gamma\sum_{h=1}^{p_{L}} \sum_{l=0}^{1} \widehat{a}_{lh}(f_{L, h}) \times \boldsymbol{1}\{f_{L, h} (\vv{x}) = l\}
\end{equation}
for each $\vv{x} \in [0, 1]^{p}$, where we recursively define $\widehat{R}_{ih} = \widehat{R}_{i, h-1} - \gamma\sum_{l=0}^{1} \widehat{a}_{lh}(f_{L, h}) \times \boldsymbol{1}\{f_{L, h} (\boldsymbol{X}_i) = l\}$ for $h\in \{1, \dots, p_{L}\}$, with $\widehat{a}_{lh}(f) = \frac{\sum_{i=1}^{n}\widehat{R}_{i, h-1} \times \boldsymbol{1}\{f (\boldsymbol{X}_{i}) = l\} }{1\vee \sum_{i=1}^{n}\boldsymbol{1}\{f (\boldsymbol{X}_{i}) = l\}}$ and $\widehat{R}_{i0} = Y_{i}$.

\subsubsection{Sample-Level Optimization of o1Neuro (Iterative Greedy Algorithm)}\label{Sec2.2.1}

\begin{figure}[ht]
  \centering
  \includegraphics[width=1.\textwidth]{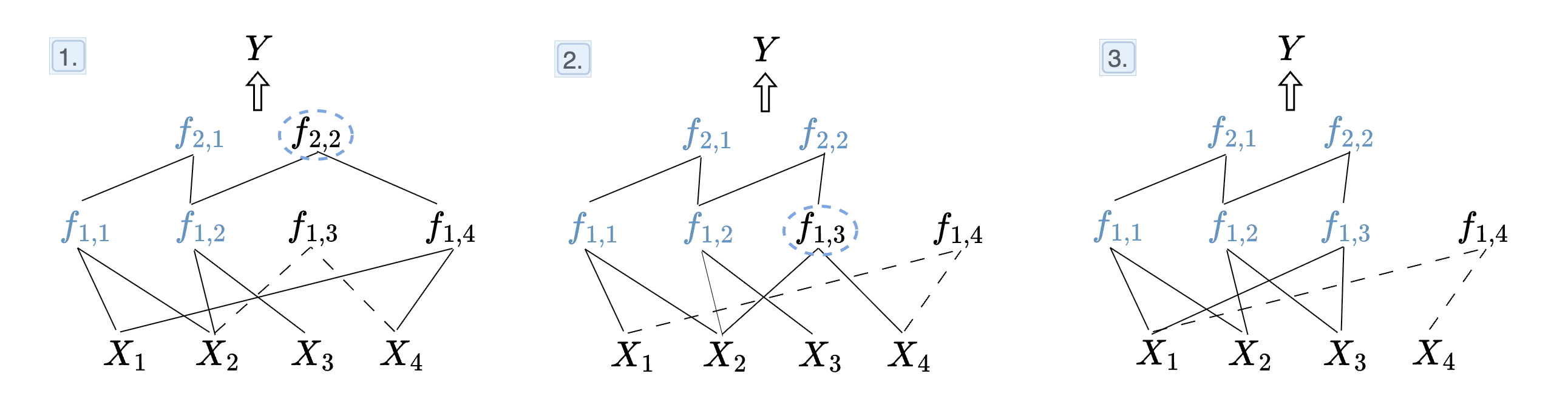}
  \caption{ Illustration of a two-hidden-layer o1Neuro network with \(L=2\), \(p_1=4\), \(p_2=2\), and  \(p_{0} = p = 4\). Solid and dashed edges indicate nonzero weights to lower neurons; dashed edges denote upper neurons unused by any output neuron. In each update round, output neurons \(f_{2,1}\) and \(f_{2,2}\) are updated sequentially. After updating \(f_{2,1}\) and its subnetwork (neurons in light blue in the left panel), \(f_{2,2}\) is next. The updated \(f_{2,2}\) assigns nonzero weights to \(f_{1,2}\) and \(f_{1,3}\), but since \(f_{1,2}\) is already updated, only \(f_{1,3}\) is updated at the first layer, completing \(f_{2,2}\)’s subnetwork update. Finally, weights and bias of the idle neuron \(f_{1,4}\) are reset by random sampling from \eqref{restriction.2}.}
  \label{fig:aro1neuro1}
\end{figure}

Sample-level optimization requires multiple update rounds (iterations) to reach convergence. In each update round, the sample-level optimization  sequentially updates each output neuron from  $f_{L, 1}$ to  $f_{L, p_L}$, as graphically illustrated in Figure~\ref{fig:aro1neuro1}. For each \(h \in \{1, \dots, p_L\}\), the subset of neurons connected to output neuron \(f_{L,h}\) is updated starting from \(f_{L,h}\) itself down to the first layer. To update $(\vv{w}_{L, h}, c_{L, h})$, we randomly sample $K$ candidate parameter pairs from \eqref{restriction.2}. Replacing $(\vv{w}_{L, h}, c_{L, h})$ with each candidate produces $K$ competitive output neurons $f_1, \dots, f_K$, along with (1) the current neuron $f_{K+1} := f_{L, h}$, and (2) $f_{K+j} := f_{L, h_j}$ for randomly sampled $h_j \in \{1, \dots, p_L\}$ with $j \in \{2, \dots, K\}$ to ensure a stable update. Each candidate is evaluated by $\widehat{W}_{h}(f) \coloneqq \sum_{l=0}^1 \frac{ \left[ \frac{1}{n}\sum_{i=1}^{n}\widehat{R}_{i, h-1} \times  \boldsymbol{1}\{f (\boldsymbol{X}_{i}) = l\} \right]^2}{ \frac{1}{n}\sum_{i=1}^{n}\boldsymbol{1}\{f(\boldsymbol{X}_{i}) = l\}}$  for every $f:\mathbb{R}^p \mapsto \{0,1\}$.
We then replace the current neuron with the candidate $s^{\star} = \argmax_{s\in \{1, \dots, 2K\}} \widehat{W}_{h}(f_{s})$ if and only if $\widehat{W}_{h}(f_{s^{\star}}) > (1 + \epsilon_0) \times \widehat{W}_{h}(f_{K + 1})$ for some small hyperparamter $\epsilon_0\ge0$. Here, choosing $\epsilon_0>0$ can accelerate and stabilize optimization. After updating $f_{L,h}$, we next update the unvisited neurons in the $(L-1)$th layer directly connected to $f_{L,h}$. Within each layer, neurons are updated sequentially in $h$. This process continues recursively to the first layer, completing the update of the subnetwork for $f_{L, h}$, as illustrated in Figure~\ref{fig:aro1neuro1}.

After updating each output neuron and its subnetwork, the remaining idle neurons are randomly refreshed with weights and biases sampled from \eqref{restriction.2}, completing one update round without incurring additional optimization cost.

\section{Theoretical Foundations of o1Neuro}\label{Sec3b}

Theorem~\ref{lemma1} (Section~\ref{Sec3.1b}) shows that, under mild conditions, deep population-level o1Neuro can approximate \textit{any measurable function} of $\boldsymbol{X}$, while a shallow o1Neuro suffices for practical additive models with complex interactions. Theorem~\ref{lemma2} (Section~\ref{Sec3.2b}) further guarantees that the sample-level model can be optimized as the sample counterpart of an optimal population-level o1Neuro, ensuring its constructive approximation capability and providing advantages over prior work reviewed in Section~\ref{Sec1.1b}. We set $M = p_{L}$ to simplify notation.


\subsection{Constructive Universal Approximation}\label{Sec3.1b}

Define 
$
\widetilde{\mathcal{G}} \coloneqq \{ \text{All possible } f_{L,1} \text{ satisfying \eqref{restriction.1}} \},
$
where “all possible” refers to all permissible choices of weights and biases. All other notations in this section are consistent with those defined in Section~\ref{Sec2.1b}.
For clarity, for each $h \in \{1, \dots, p_{L}\}$, we have
$
\widetilde{\mathcal{G}} = \{ \text{All possible } f_{L,h} \text{ satisfying \eqref{restriction.1}} \}.
$
Condition~\ref{condi.inde} is a distributional assumption on $(Y, \boldsymbol{X})$.

\begin{condition}\label{condi.inde}
The distribution of $\boldsymbol{X}$ has a bounded density. In addition, the function $\mathbb{E}[ Y\mid \boldsymbol{X} = \vv{x}]$ belongs to
$\mathcal{L}(k, R_0)\coloneqq \{\sum_{r=1}^{R_0} f_r(\vv{x}): f_r\in \mathcal{L}(k)\}$ for some integer $R_0>0$, where
$\mathcal{L}(k)  = \{ f : [0, 1]^{p} \mapsto \mathbb{R} \mid  \mathbb{E}[f(\boldsymbol{X})]^2<\infty \text{ and }  f(\boldsymbol{X}) = g(X_{i_1}, \dots, X_{i_k})  
             \textnormal{ for some  }  
			g : [0, 1]^{k} \mapsto \mathbb{R} \textnormal{ and }  \{i_{1}, \dots, i_{k}\} \subset \{1, \dots, p\}\}.$
\end{condition}

\begin{theorem}\label{lemma1}
Let an arbitrary constant $t \in(0, 1]$ be given. Let a population optimal o1Neuro, as defined in Section~\ref{Sec2.1.1}, have output neurons $f_{L,1}, \dots, f_{L,M}$, with $p_l \ge 2^{L-l} M$ for each $l \in \{1, \dots, L-1\}$.
    Then, for each $h\in \{1, \dots, p_{L}\}$, 
    \begin{equation}
\begin{split}
    \label{eq.1}
    & \sum_{l=0}^1  \left|\mathbb{E}[\widetilde{R}_{h-1} \times \frac{\boldsymbol{1}\{f_{L,h} (\boldsymbol{X}) = l\}}{\sqrt{\mathbb{P}(f_{L,h} (\boldsymbol{X}) = l) }} ] \right|^2   \ge t\times \sup_{g\in\widetilde{\mathcal{G}}} \sum_{l=0}^1 \left|\mathbb{E}[\widetilde{R}_{h-1} \times \frac{\boldsymbol{1}\{g (\boldsymbol{X}) = l\}}{\sqrt{\mathbb{P}(g (\boldsymbol{X}) = l)} } ] \right|^2,
    \end{split}
\end{equation}
implying $\lim_{M \to \infty} \mathbb{E}\!\left[\widetilde{m}(\boldsymbol{X}) - \mathbb{E}[Y \mid \boldsymbol{X}]\right]^2 = 0$ if Condition~\ref{condi.inde} also holds with $L \ge 2 + \lceil\log_2 k\rceil$.
     \end{theorem}

\begin{figure}[htbp]
    \centering    
        \includegraphics[width=1.\textwidth]{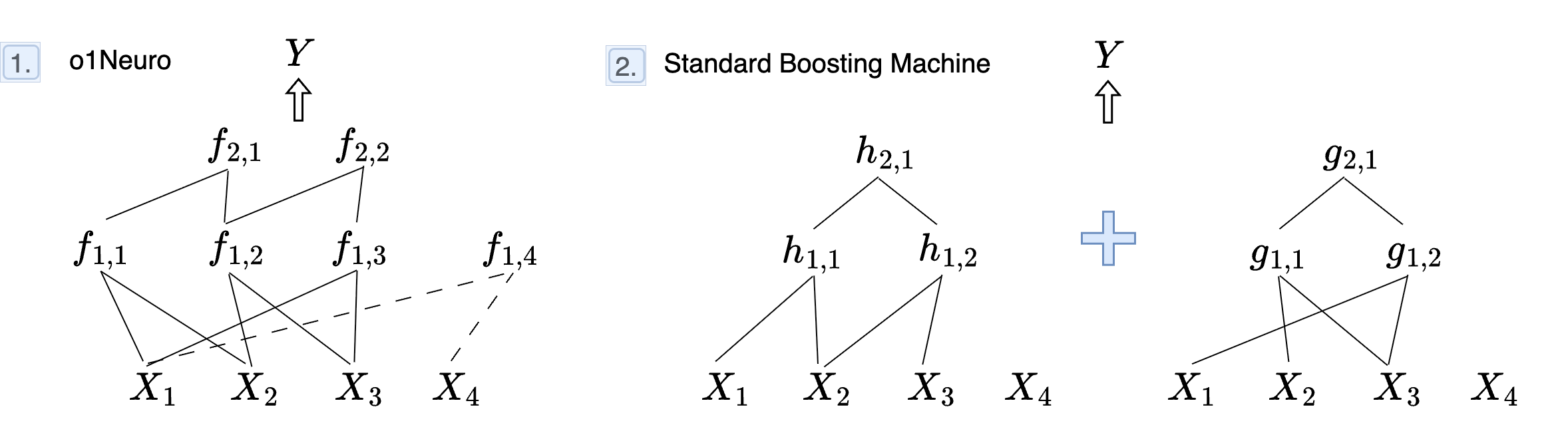}    
\caption{In the second panel, both $g_{2,1}$ and $h_{2,1}$ are from $\widetilde{\mathcal{G}}$ at $L=2$. The networks in both panels are equivalent at the population level, but each $f_{2,h}$ of o1Neuro can take any of $f_{1,1},\dots,f_{1,4}$ as input, enabling more efficient sample-level optimization.}
    \label{fig:compa}
\end{figure}

With $t=1$, Theorem~\ref{lemma1} shows that our population o1Neuro behaves like a standard boosting machine~\citep{friedman2001greedy}, where an optimal function from $\widetilde{\mathcal{G}}$ is iteratively added to the predictive model. A standard boosting predictor is illustrated in the second panel of Figure~\ref{fig:compa}. In contrast, o1Neuro allows each neuron $f_{l,h}$ to take inputs from any neuron in the $(l-1)$th layer, enabling more efficient sample-level optimization. We will see in Theorem~\ref{lemma2} that the sample optimal o1Neuro satisfies the sample analogue of \eqref{eq.1}.

The second part of Theorem~\ref{lemma1} shows that a properly tuned o1Neuro can approximate a broad class of data-generating functions in $\mathcal{L}(k, R_0)$ with $k \le p$, including cases where $\mathbb{E}(Y\mid \boldsymbol{X}) = h(X_1, \dots, X_k)$ for any measurable $h:[0,1]^k \mapsto \mathbb{R}$. For example, when $L = 2 + \lceil \log_2 p \rceil$, o1Neuro can approximate \textit{any measurable function} on $[0,1]^p$. When $L = 3$, it can approximate any additive model of the form  
$
\mathbb{E}(Y \mid \boldsymbol{X}) = \sum_{r=1}^{R_0} h_r(X_{2r-1}, X_{2r})
$
for any measurable $h_r : [0, 1]^2 \mapsto\mathbb{R}$, which may include multiple XOR interaction components as well as univariate functions.

These results have two important implications.  
First, from a theoretical perspective, Theorem~3.8 of \citep{barron2008approximation} similarly shows that a one-hidden-layer \emph{non-sparse} neural network, constructed via an orthogonal (or relaxed) greedy algorithm, can approximate any measurable function of $\boldsymbol{X}$. Taken together, our results and theirs reveal a trade-off between activation sparsity and network depth in achieving strong approximation capability.

Second, from a practical perspective, Theorem~\ref{lemma1} demonstrates that o1Neuro maintains strong approximation power even when employing sparse activations~\citep{lee1996efficient} to mitigate the curse of dimensionality and shallow networks to accelerate optimization. In particular, shallow o1Neuro configurations can still approximate a wide range of commonly encountered functions, including additive models~\citep{friedman2001greedy} and interaction effects~\citep{bien2013lasso, cox1984interaction}, making them sufficient for most prediction tasks. This shows that these practical design choices do not compromise approximation capability. By comparison, most existing approximation theories~\citep{schmidt2020nonparametric, bauer2019deep, yarotsky2018optimal, jiao2023deep, barron2008approximation} focus on universal approximation while largely ignoring such practical optimization considerations.

It should be noted that our predictor \eqref{population.01} is a boosted model~\citep{friedman2001greedy} that selects base learners from $\widetilde{\mathcal{G}}$, whereas Theorem~\ref{lemma1} is derived from greedy approximation theory~\citep{temlyakov2000weak}.
 To our knowledge, Theorem~\ref{lemma1} is the first result to formally analyze a boosting machine using greedy approximation theory, and it also represents the first application of boosting machine to deep neural networks. Although \eqref{population.01} satisfies \eqref{eq.1} with $t=1$, we retain $t \in (0,1]$ for consistency with the literature, where $t<1$ represents a weak greedy approximation that accounts for statistical estimation and optimization stability \citep{temlyakov2000weak}. See Section~\ref{Sec2.2.1} for the choice of $t= \frac{1}{1 + \epsilon_0} > 0$ to ensure optimization stability.

\begin{remark}
Based on the discussion in Section~4.1 of \citep{barron2008approximation}, Theorem~\ref{lemma1} may also hold with $L = 1$ when $k = 2$. It may be possible to refine Theorem~\ref{lemma1} further to obtain a tighter lower bound for $L$. Additionally, comparing the approximation capabilities and optimization efficiency of single-hidden-layer non-sparse networks~\citep{barron2008approximation} with multi-layer sparse networks (o1Neuro) is of interest. We leave these potential refinements and discussions for future work.
\end{remark}

\begin{remark}
    Consistency of boosting machines has been studied in recent work~\citep{biau2021optimization}, but that study does not specify the class of target functions as in Condition~\ref{condi.inde}. A goal of Theorem~\ref{lemma1} is to establish that o1Neuro can serve as a universal approximator for any measurable data-generating function given proper network depth.
\end{remark}

\begin{remark}
Since our current approximation theory does not provide convergence rates, we cannot yet analyze, based on Theorem~\ref{lemma1}, how o1Neuro mitigates the curse of dimensionality or how the boosting learning rate $\gamma$ influences approximation convergence. Dimensionality-free approximation convergence rates have been studied in \citep{jiao2023deep, barron1992neural} and references therein. However, to the best of our knowledge, existing dimensionality-free approximation theories are all nonconstructive.


\end{remark}

\subsection{Sure Convergence}\label{Sec3.2b}
Theorem~\ref{lemma2} below shows that the optimization of our sample o1Neuro converges after sufficient updates, after which the model parameters remain unchanged. In Theorem~\ref{lemma2}, each $f_{L,h}$ denotes the output neuron of a trained o1Neuro network as in \eqref{sample.predictor.1} after $b$ update rounds (see Section~\ref{Sec2.2.1}), assuming 
 \begin{equation}
     \label{sample.p}
     p_{l} \ge 2^{L-l} M+ 2^{L-l}
 \end{equation}
 for $l \in \{1, \dots, L-1\}$.  Define the sample analogue of \(\widetilde{\mathcal{G}}\) as 
 $$\widehat{\mathcal{G}} \coloneqq \{ \text{All possible } f_{L,1} \text{ satisfying \eqref{restriction.2}} \}.$$

The additional $2^{L-l}$ neurons in~\eqref{sample.p} at each $l$th layer during sample-level optimization (Theorem~\ref{lemma1} requires $p_{l} \ge 2^{L-l} M$ instead) provide sufficient capacity for a subnetwork of idle neurons corresponding to the size of functions in $\widehat{\mathcal{G}}$. As shown in the proof of Theorem~\ref{lemma2} in Section~\ref{proof.theorem2}, this subnetwork can realize any function in $\widehat{\mathcal{G}}$ with positive probability when randomly refreshed each round, due to the zero-gradient regions induced by the indicator neuron functions. Specifically, the zero-gradient property means that, for a fixed sample, the function mapping weights and biases to the o1Neuro sample loss is a simple function. This property, together with idle neurons and iterative optimization, underlies Theorem~\ref{lemma2}. It should be noted that we initialize all model parameters to zero, without requiring any specific parameter settings (see Section~\ref{Sec2.2b});  all other notations follow the same section.

\begin{theorem}
\label{lemma2}
Let $K > 0$ be an integer and $\epsilon_0 \ge 0$ a real constant. As $b \to \infty$, with probability approaching one, the optimization completes such that for each $h \in \{1, \dots, p_{L}\}$,
{\small\begin{equation}
    \label{eq.2}
     \sum_{l=0}^1 \frac{ \left[ \frac{1}{n}\sum_{i=1}^{n}\widehat{R}_{i, h-1}  \boldsymbol{1}\{f_{L, h} (\boldsymbol{X}_{i}) = l\} \right]^2}{ \frac{1}{n}\sum_{i=1}^{n}\boldsymbol{1}\{f_{L, h}(\boldsymbol{X}_{i}) = l\}} \ge \frac{1}{1 + \epsilon_0} \max_{g\in \widehat{\mathcal{G}} }\sum_{l=0}^1 \frac{ \left[ \frac{1}{n}\sum_{i=1}^{n}\widehat{R}_{i, h-1}  \boldsymbol{1}\{g (\boldsymbol{X}_{i}) = l\} \right]^2}{\frac{1}{n}\sum_{i=1}^{n}\boldsymbol{1}\{g (\boldsymbol{X}_{i}) = l\}}.
\end{equation}}
 
\end{theorem}

Theorem~\ref{lemma2} shows that, after a sufficient number of update rounds, $\widehat{m}$ in \eqref{sample.predictor.1} reaches an optimal solution (i.e., optimization convergence) with probability approaching one and satisfies \eqref{eq.2}, serving as the sample-level analogue of \eqref{eq.1}.
 Together, Theorem~\ref{lemma2} and Theorem~\ref{lemma1} establish that o1Neuro enjoys both optimization convergence and universal approximation guarantees under a reasonable setup. These two properties distinguish o1Neuro from existing approaches reviewed in Section~\ref{Sec1.1b}. In what follows, we discuss the required lower bound on $b$ for completing optimization in our iterative greedy algorithm.

Example~\ref{example1} complements Theorem~\ref{lemma2} by showing that the lower bound on $b$ is explicitly analyzable and remains well-bounded in certain cases. All parameters are fixed except for $n$ and $\min_{1 \le l \le L-1} p_l$, with the proof in Section~\ref{proof.example1}.
\begin{exmp}\label{example1}
     Assume $(\boldsymbol{X}_1, Y_1), \dots, (\boldsymbol{X}_n, Y_n), (\boldsymbol{X}, Y)$ are i.i.d., where 
$Y = \sum_{j=1}^{R_0} \beta_j X_{j}$ and 
$2^{-p+1} \beta_j^2 > \sum_{l=j+1}^{R_0} \beta_{l}^2$ for each $j \in \{1, \dots, R_0\}$. 
Here $\boldsymbol{X}$ is uniformly distributed on $\{0, 1\}^p$, $p \ge R_0$. 
Consider a sample o1Neuro network of depth $L$ with $p_{L} = R_0$, updated for $b$ rounds with $\gamma = 1$ and $\epsilon_0 = 0$. 
If $b \ge 4^{L} \kappa R_0 p$ for some $\kappa >0$, then \eqref{eq.2} holds with probability at least $1 - R_0 e^{-\kappa K}$, provided $n$ and $\min_{1 \le l \le L-1} p_l$ are sufficiently large.
\end{exmp}

 The number of hidden layers $L$ of o1Neuro only needs to be moderate and finite, as discussed after Theorem~\ref{lemma1}. Setting $K = p$ matches the number of orthogonal split candidates per node in tree models~\citep{breiman2001random} with binary inputs. In the configuration with fixed $L$, $K = p$, and $\kappa = c_0 p^{-1}$ for some $c_0 \ge 1$, Example~\ref{example1} complements Theorem~\ref{lemma2} by showing that the computational requirement on $b$ grows at most linearly with $R_0$, the number of additive components in the data-generating function.

The distributional assumptions on $(\boldsymbol{X}, Y)$ in Example~\ref{example1} are limited to a noiseless linear model with exponentially decaying coefficients and binary features, which simplifies the derivations. For more complex data-generating functions, the required lower bound on $b$ has been empirically verified as manageable (e.g., $b\le 5$) for high-quality o1Neuro models in Sections~\ref{Sec6b}--\ref{Sec7b}, with $K$ fixed at 15.

On the other hand, our neural network optimization is rarely carried out to completion in practice. Theorem~\ref{lemma2} ensures that the sample o1Neuro still effectively serves as a proxy for a model satisfying \eqref{eq.2}. Empirically, high accuracy of o1Neuro is often reached within just a few updates, which may relate to the observation that locally optimal models can perform comparably to globally optimal ones in standard gradient-based neural networks~\citep{choromanska2015loss}. A theoretical explanation in the context of greedy algorithms remains open.




\section{Hyperparameter Optimization}\label{Sec5b}

To accelerate hyperparameter optimization, we employ stochastic training, in which each update round is performed on a randomly sampled subset of the training data of size $\textsf{stochastic\_ratio} \times$ (total training samples) for some $\textsf{stochastic\_ratio} \in (0,1]$, and a neuron freezing strategy, where a random subset of neurons, corresponding to $\textsf{freezing\_rate} \times$ (total number of neurons) with $\textsf{freezing\_rate} \in [0,1)$, is temporarily frozen. Freezing strategies are commonly used to accelerate network training~\citep{brock2017freezeout}, often in different forms. Both techniques are applied only during hyperparameter optimization to reduce computational cost.

\subsection{Training Schedule}\label{Sec5.1b}
For all models, we use the Python package \pkg{hyperopt} to optimize hyperparameters over $30$ trials, each time splitting the full training data into $80\%$ for training and $20\%$ for validation. After hyperparameter optimization, the final predictive model is retrained using the full training dataset. The hyperparameter spaces for o1Neuro are listed in Table~\ref{tab:validation_space_o1neuro} and briefly justified in Section~\ref{Sec5.4b}, while those for XGBoost~\citep{chen2016xgboost}, Random Forests~\citep{breiman2001random}, and TabNet~\citep{arik2021tabnet} are detailed in Section~\ref{SecA}.

During the tuning process, we set $\textsf{stochastic\_ratio} = 0.05$ and $\textsf{freezing\_rate} = 0.6$ to accelerate hyperparameter optimization. For formal model training, stochastic training and neuron freezing are not used. Additional details are provided in the caption of Table~\ref{tab:validation_space_o1neuro}. All o1Neuro architectures considered in Table~\ref{tab:validation_space_o1neuro} satisfy condition \eqref{sample.p}.

\begin{table}[htbp]
\centering
\caption{
We set $K = 15$, $\epsilon_0 = 0.01$, and $b = 5$ (see Section~\ref{Sec2.2.1} for notation). For $\text{n\_layer} = 2$, the first and second layers contain $\text{n\_neurons}$ and $2\times(\text{n\_neurons}+1)$ neurons, respectively; for $\text{n\_layer} = 1$, the first layer has $\text{n\_neurons}$ neurons.
}
\label{tab:validation_space_o1neuro}
\begin{tabular}{ll}
\hline
Parameter name & o1Neuro Search Space \\
\hline
$\gamma$ (learning rate) & Uniform $(0.0, 0.6]$ \\
$\text{n\_layer}$ & $\{1, 2\}$ \\
$\text{n\_neurons}$ & $\text{Uniform}\{450 + 350 \times (2 - \text{n\_layer}), \dots, 650 + 350 \times (2 - \text{n\_layer})\}$ \\
\hline
\end{tabular}
\end{table}

\section{Simulation Study} \label{Sec6b}

In this section, our objectives are to demonstrate o1Neuro's strong approximation capability, show that it can be optimized within only a few rounds generally, and justify our choice of hyperparameter space along with its computational efficiency. The prediction performance of o1Neuro is compared with Random Forests~\citep{breiman2001random}, a widely used bagging tree model; XGBoost~\citep{chen2016xgboost}, a well-established additive tree boosting method recommended by \citep{grinsztajn2022tree}; and TabNet~\citep{arik2021tabnet}, a recent deep learning model for tabular data.

After performing hyperparameter optimization as described in Section~\ref{Sec5.1b}, we evaluate the final predictive model using the R$^2$ score on an independently generated, error-free test set of size $10{,}000$, where R$^2$ is defined as
\begin{equation}
\label{r_square_score}
\text{R}^2 = 1 - \frac{\text{sum of squared residuals}}{\text{sample variance of the responses}}.
\end{equation}

\subsection{Data-Generating Models for Experiments}

We consider the following data-generating models:
\begin{align}
    Y &= \sum_{j=1}^{10} 2 X_j + \varepsilon, & \text{(linear model)} \label{Y1}\\
    Y &= \sum_{j=1}^{10} 2\, X_{2j-1} X_{2j} + \varepsilon, & \text{(additive model with XOR interactions)} \label{Y2}
\end{align}
where $\boldsymbol{X} = (X_{1}, \dots, X_{20})^{\top}$ are independently sampled from a uniform distribution on $[-0.5, 0.5]^{20}$, and $\varepsilon \sim \mathcal{N}(0,1)$ is an independent noise term. We set the sample size to $n \in \{100, 500, 3000, 20000\}$ to enable a fair comparison of predictive performance, and each experiment is repeated 10 times to ensure stable results.

The models \eqref{Y1}--\eqref{Y2} are chosen for two reasons. Model \eqref{Y2} includes XOR-type interactions, which are challenging for simple one-hidden-layer networks~\citep{elman1990finding} and common in practice~\citep{bien2013lasso, cox1984interaction}. Both models are additive, allowing a clearer comparison between o1Neuro and boosting methods (XGBoost), which are especially effective for such structures.

\subsection{Results}

Table~\ref{tab:prediction_comparison} summarizes the predictive performance of the models under different data-generating processes and sample sizes. For the linear model~\eqref{Y1}, both o1Neuro and XGBoost achieve consistently high R$^2$ values, with o1Neuro performing on par with XGBoost, while Random Forests lag behind, underscoring the advantage of the boosting framework.
In the additive XOR model~\eqref{Y2}, o1Neuro outperforms all other methods, and the performance gap persists as the sample size increases, demonstrating its strong capability to approximate complex functions. XGBoost remains competitive in this setting, whereas Random Forests deliver moderate performance. TabNet achieves reasonable performance on larger sample sizes but shows lower accuracy on smaller samples, reflecting its sensitivity to data size and the importance of careful hyperparameter tuning. Despite the extensive universal approximation theory for deep learning and TabNet's design for tabular data regression, its modest performance, particularly on small samples, indicates a gap between theoretical guarantees and practical effectiveness, as noted in the literature~\citep{adcock2021gap, grohs2024proof}.

\begin{table}[ht]
\centering
\begin{tabular}{lcccc}
\hline
 Model/$n$ & \eqref{Y1} / $100$ & \eqref{Y1} / $500$ & \eqref{Y2} / $3000$ & \eqref{Y2} / $20000$ \\
\hline
o1Neuro        &   0.694 (0.079)   &   0.890 (0.03)   &    0.544 (0.102)	& 0.879 (0.031)   \\
TabNet         &    -0.099 (0.095)   &   0.255 (0.241)   &   -0.256 (0.191)   &   0.282 (0.212)   \\
XGBoost        &   0.684 (0.067)   &   0.897 (0.028)   &   0.405 (0.118)   & 0.744 (0.090)      \\
Random Forests &   0.417 (0.074)   &   0.670 (0.021)   &   0.206 (0.02)   &  0.356 (0.006)    \\
\hline
\end{tabular}
\caption{Mean prediction R$^2$ scores (standard deviation in parentheses) across models and sample sizes over 10 repetitions.}

\label{tab:prediction_comparison}
\end{table}

\subsubsection{Optimized Learning Rates for o1Neuro}

The boosting learning rate $\gamma \in (0,1]$ is the most critical hyperparameter for o1Neuro's generalization. In particular, while overfitting may occur in artificial data (se  Figure~\ref{fig:r2_grid}), restricting $\gamma$ to $(0,0.6]$ in practice (Table~\ref{tab:validation_space_o1neuro}) effectively prevents noticeable overfitting in real data (Section~\ref{Sec7b}). Moreover, when computational cost is less critical, full models without stochastic training or neuron freezing can be used, which allows better generalization without constraining $\gamma$. On the other hand, regarding architecture, the two-hidden-layer configuration is selected more often for the linear model~\eqref{Y1} (approximately half of the time across $10$ trials) than for the additive XOR model~\eqref{Y2}, for which a single hidden layer, known to be the simplest network architecture for handling XOR-type interactions~\citep{elman1990finding}, is almost always preferred. A plausible explanation is that o1Neuro favors simpler network architectures when given sufficient training data. This also suggests that the $L=3$ requirement (number of hidden layers) in Theorem~\ref{lemma1} for approximating functions in $\mathcal{L}(2)$ may be overly conservative, with $L=1$ appearing sufficient.

\subsubsection{Prediction Performance of o1Neuro Across Update Rounds}\label{Sec6.3.2}

\begin{figure}[htbp]
    \centering
    \begin{tabular}{cc}
        \includegraphics[width=0.45\textwidth]{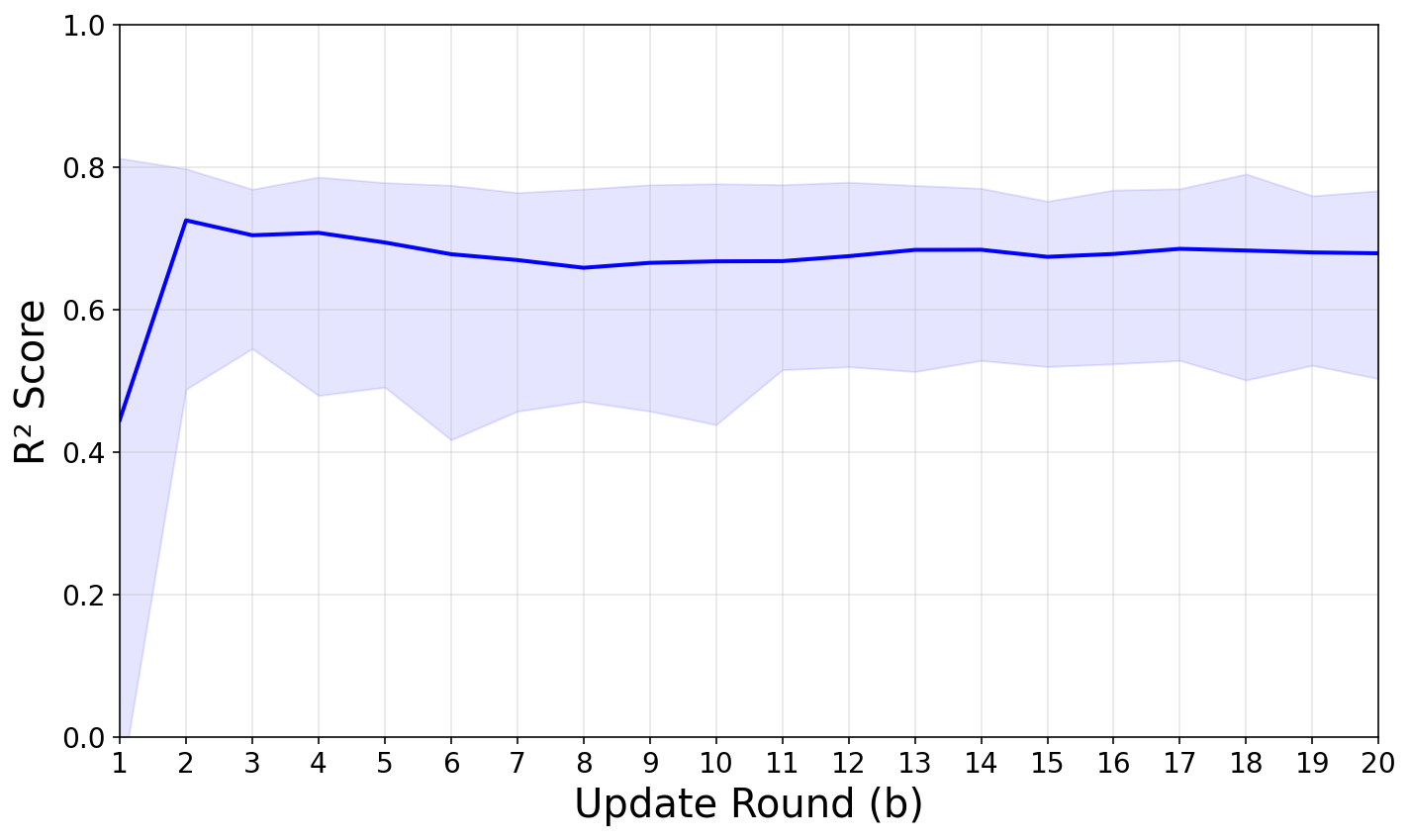} &
        \includegraphics[width=0.45\textwidth]{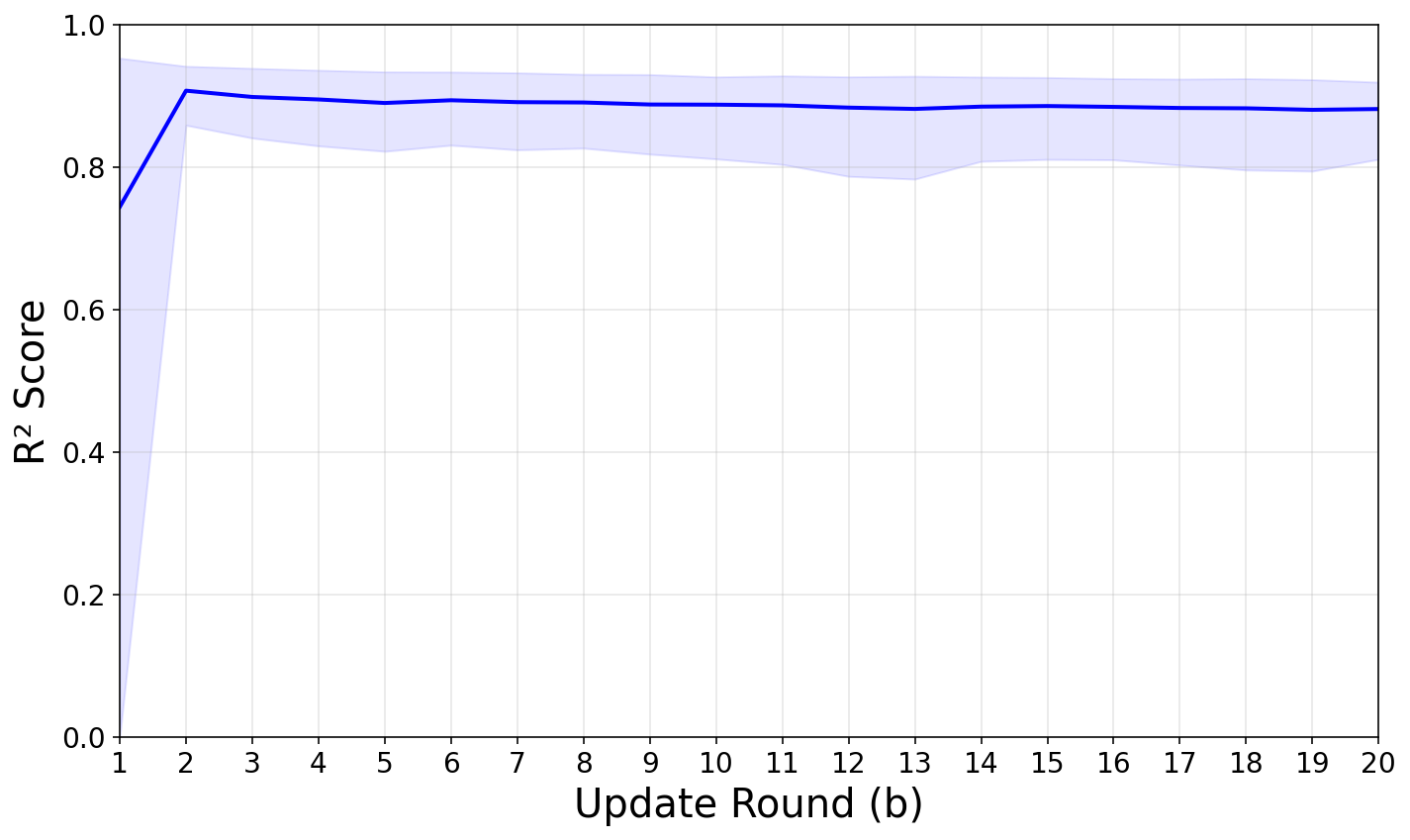} \\
        \includegraphics[width=0.45\textwidth]{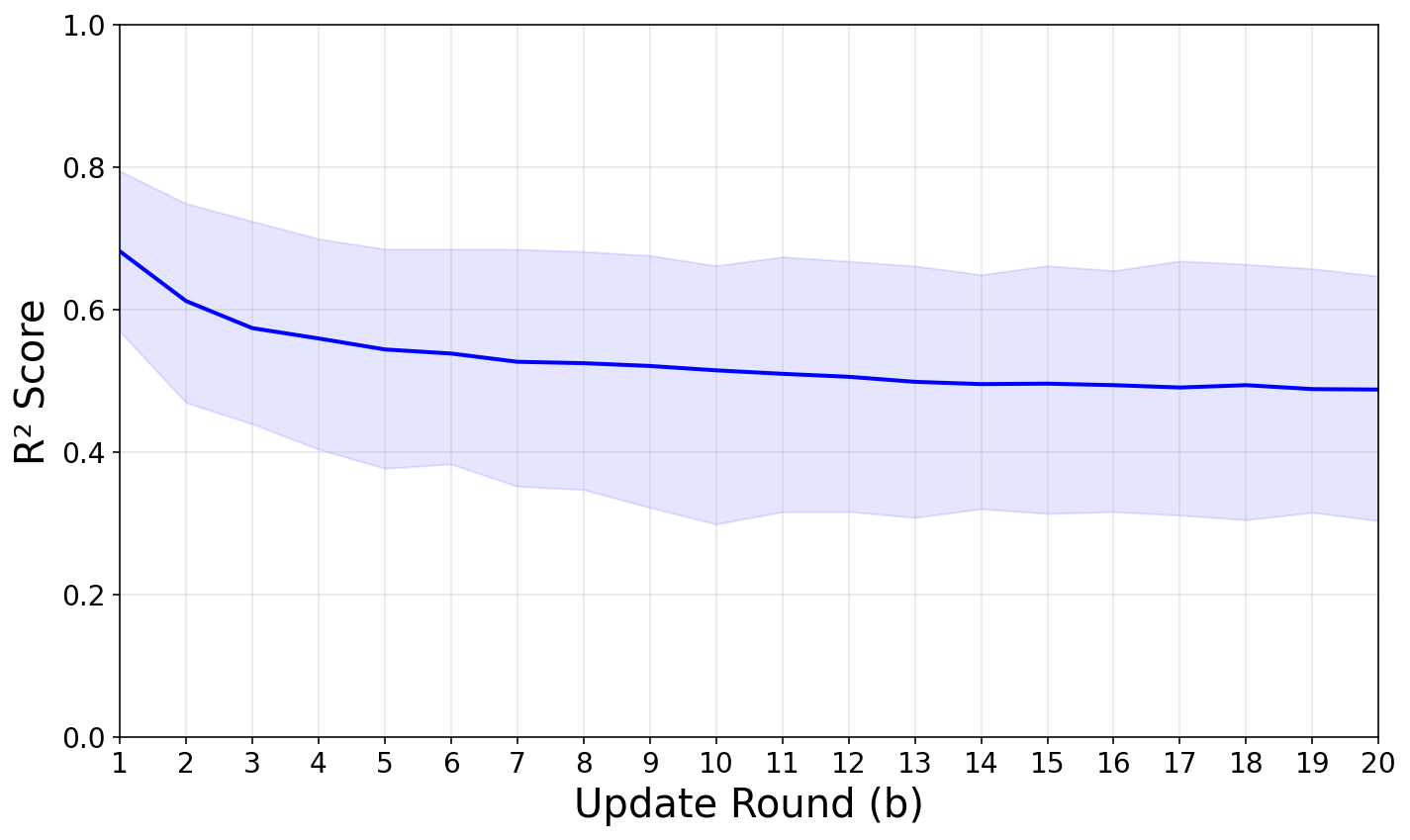} &
        \includegraphics[width=0.45\textwidth]{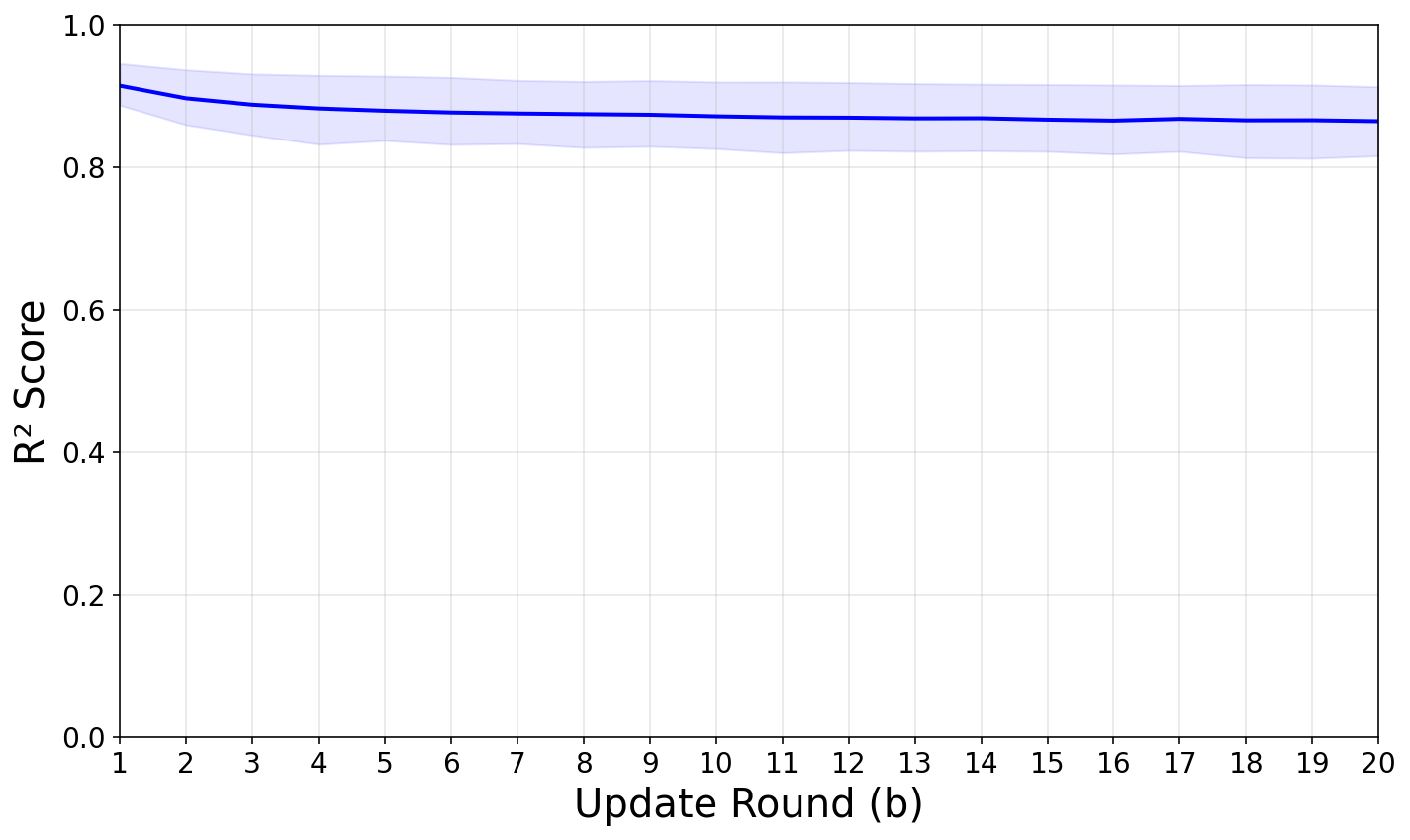} \\
    \end{tabular}
\caption{R$^2$ of o1Neuro across $b \in \{1, \dots, 20\}$; Table~\ref{tab:prediction_comparison} shows $b = 5$. Top row: Model \eqref{Y1} with $n = 100$ (left) and $n = 500$ (right). Bottom row: Model \eqref{Y2} with $n = 3000$ (left) and $n = 20000$ (right). Shaded areas show min–max range, solid lines show the mean.}

    \label{fig:r2_grid}
\end{figure}

Under both models \eqref{Y1}--\eqref{Y2}, the R$^2$ curves (based on test samples) in Figures~\ref{fig:r2_grid} initially rise  in the first or second update round and then decline as the number of update rounds increases, indicating mild overfitting. This effect is less pronounced under the linear model~\eqref{Y1} or when larger training samples are available. With small sample sizes or models that include complex interactions, the R$^2$ curves exhibit greater performance variability; however, increasing the sample size substantially stabilizes o1Neuro's performance across both modeling scenarios. Notably, the jumps at the second update round in the top two R$^2$ curves of Figure~\ref{fig:r2_grid} arise from using two-hidden-layer networks, as zero-initialized parameters often lead to poor initial performance in deeper architectures.

\subsubsection{Computing Runtime}

\begin{table}[ht]
\centering
\begin{tabular}{lcccc}
\hline
  & \multicolumn{2}{c}{Seconds Per Configuration} & \multicolumn{2}{c}{Seconds Per Predictive Model} \\
\hline
  & $n=3{,}000$ & $n=20{,}000$ & $n=3{,}000$ & $n=20{,}000$ \\
\hline
o1Neuro        &   6.8   &  18.7    &   53.5   &    715  \\
TabNet         &    121  &  1251    &  121    &  1251   \\
XGBoost        &   2.4   &  7.8    &   2.4   &    7.8  \\
Random Forests &   5.6  &   214   &    5.6  &    214  \\
\hline
\end{tabular}
\caption{Computing runtime for each model under the data-generating model \eqref{Y2}. Training a predictive o1Neuro model involves five update rounds ($b=5$), whereas for the other models, the required time per configuration and per predictive model is the same. Experiments were conducted on a 2023 Mac Studio with an Apple M2 Ultra chip, featuring a 24-core CPU, 60-core GPU, and 64~GB memory.
}
\label{tab:runtime}
\end{table}

By design, the computational runtime of o1Neuro is expected to scale approximately linearly with the training sample size, though small deviations in Table~\ref{tab:runtime} may arise from hardware overhead, hyperparameter choices, or limited test repetitions. Training a predictive o1Neuro model takes roughly 700 seconds for $n = 20{,}000$, but runtime can be reduced using stochastic training and neuron freezing as in Section~\ref{Sec5b}; we do not consider this accelerated version for o1Neuro prediction, since the total cost of approximately $(30 \times 18.7 + 715)$ seconds $\approx 21$ minutes is reasonable when $n=20{,}000$, and the goal is to evaluate the prediction performance of the vanilla o1Neuro. The accelerated version is used only for hyperparameter optimization. XGBoost scales efficiently with sample size, whereas Random Forests are less stable, likely due to hyperparameter sensitivity, with full optimization and training taking about $(30 \times 214 + 214)$ seconds $\approx 107$ minutes for $n = 20{,}000$. TabNet requires at least 1000 seconds per configuration, making it impractical for large samples.

\subsubsection{Architecture Selection}\label{Sec5.4b}

The o1Neuro architecture space recommended in Table~\ref{tab:validation_space_o1neuro} demonstrates strong performance for tabular data regression in our experiments. While shallow networks often require many neurons to approximate complex functions~\citep{telgarsky2016benefits}, our results show that even for the sophisticated model~\eqref{Y2}, the proposed shallow o1Neuro architectures achieve competitive performance comparable to boosting methods such as XGBoost, which considers up to 1000 boosted trees in its standard hyperparameter space (Section~\ref{SecA}). Increasing the depth of o1Neuro yields only marginal improvements that do not justify the additional runtime in our experiments. Although our empirical study focuses on shallow architectures, motivated by their strong practical performance rather than subjective preference, we note based on additional unreported experiments that for deeper o1Neuro networks the requirement in~\eqref{sample.p} is generally unnecessary in practice.

\section{Prediction Evaluation on Real Datasets}\label{Sec7b}


\begin{table}[htbp]
\centering
\setlength{\tabcolsep}{6pt} 
\caption{R$^2$ scores and ranks on four datasets over 10 repetitions, with $p$ features and $n_{\text{train}}$ training samples. Left: original $p$ features; Right: transformed $2p$ features.}
\label{tab:r2_results_ranked_ties_compact}
\begin{minipage}{0.48\textwidth}
\centering
{\footnotesize
\begin{tabular}{lcccc}
\hline
& Max & Mean (Std) & Min & Rank \\
\hline
\multicolumn{5}{c}{elevators ($p = 16$, $n_{\text{train}} = 6000$)} \\
o1Neuro       & 0.881 & 0.874 (0.004) & 0.867 & 2 \\
TabNet        & 0.329 & -0.407 (0.601) & -1.338 & 4 \\
XGBoost       & 0.896 & 0.882 (0.008) & 0.868 & 1 \\
RF            & 0.812 & 0.764 (0.027) & 0.705 & 3 \\
\hline
\multicolumn{5}{c}{Ailerons ($p = 33$, $n_{\text{train}} = 6000$)} \\
o1Neuro       & 0.853 & 0.840 (0.008) & 0.824 & 1 \\
TabNet        & -0.609 & $\ll 0$ & $\ll 0$ & 4 \\
XGBoost       & 0.832 & 0.825 (0.006) & 0.817 & 3 \\
RF            & 0.836 & 0.826 (0.006) & 0.815 & 2 \\
\hline
\multicolumn{5}{c}{medical\_charges ($p = 3$, $n_{\text{train}} = 6000$)} \\
o1Neuro       & 0.981 & 0.977 (0.002) & 0.973 & 1 \\
TabNet        & 0.968 & 0.567 (0.324) & -0.049 & 4 \\
XGBoost       & 0.981 & 0.977 (0.003) & 0.973 & 1 \\
RF            & 0.980 & 0.977 (0.003) & 0.972 & 1 \\
\hline
\multicolumn{5}{c}{abalone ($p = 7$, $n_{\text{train}} = 2506$)} \\
o1Neuro       & 0.577 & 0.544 (0.022) & 0.500 & 2 \\
TabNet        & 0.569 & 0.500 (0.053) & 0.421 & 3 \\
XGBoost       & 0.559 & 0.492 (0.041) & 0.416 & 4 \\
RF            & 0.560 & 0.545 (0.010) & 0.527 & 1 \\
\hline
\end{tabular}}
\end{minipage}
\hfill
\begin{minipage}{0.48\textwidth}
\centering
{\footnotesize
\begin{tabular}{ccccc}
\hline
& Max & Mean (Std) & Min & Rank \\
\hline
\multicolumn{5}{c}{elevators ($p = 32$, $n_{\text{train}} = 6000$)} \\
 & 0.839 & \textbf{0.821} (0.011) & 0.801 & 1 \\
 & 0.056 & -0.914 (0.611) & -2.202 & 4 \\
 & 0.743 & 0.705 (0.021) & 0.663 & 2 \\
 & 0.568 & 0.539 (0.019) & 0.506 & 3 \\
\hline
\multicolumn{5}{c}{Ailerons ($p = 66$, $n_{\text{train}} = 6000$)} \\
 & 0.815 & \textbf{0.801} (0.008) & 0.784 & 1 \\
 & -0.725 & $\ll 0$ & $\ll 0$ & 4 \\
 & 0.687 & 0.669 (0.011) & 0.652 & 3 \\
 & 0.687 & 0.673 (0.009) & 0.657 & 2 \\
\hline
\multicolumn{5}{c}{medical\_charges ($p = 6$, $n_{\text{train}} = 6000$)} \\
 & 0.980 & 0.977 (0.002) & 0.974 & 1 \\
 & 0.974 & 0.344 (0.503) & -0.724 & 4 \\
 & 0.980 & 0.977 (0.002) & 0.974 & 1 \\
 & 0.980 & 0.972 (0.005) & 0.960 & 3 \\
\hline
\multicolumn{5}{c}{abalone ($p = 14$, $n_{\text{train}} = 2506$)} \\
 & 0.529 & 0.503 (0.016) & 0.469 & 2 \\
 & 0.449 & 0.306 (0.110) & 0.165 & 4 \\
 & 0.492 & 0.492 (0.026) & 0.439 & 3 \\
 & 0.524 & 0.505 (0.011) & 0.488 & 1 \\
\hline
\end{tabular}}
\end{minipage}
\end{table}

In this section, we evaluate the ability of o1Neuro to capture complex effects in real applications. Since most real datasets lack known data-generating functions, we apply a minimal feature transformation to ensure interaction components. Each input feature \(X_j\) is replaced by two features \((X_{j}U_{j}^{-1}, U_{j})\), where \(U_j\) is an independent Rademacher random variable, i.e., \(\mathbb{P}(U_j = -1) = \mathbb{P}(U_j = 1) = 0.5\). We restrict our focus to two-way interactions, as they are more common in real applications~\citep{bien2013lasso, cox1984interaction}.

The benchmark models are the same as those in Section~\ref{Sec6b}. For a fair comparison, we select four datasets: elevators, Ailerons, medical\_charges, and abalone, from OpenML~\citep{OpenML2013} and the UCI Machine Learning Repository~\citep{Dua:2017}. These datasets are chosen because benchmark models are known to perform well on them~\citep{gentile2024approximation, mcelfresh2023neural}. We use the R\(^2\) measure and limit each dataset to at most \(n_{\text{dataset}} = 10{,}000\) samples, randomly subsampling when necessary. The evaluation methodology follows established procedures in~\citep{mcelfresh2023neural, grinsztajn2022tree}.

The R\(^2\) score in~\eqref{r_square_score} is computed using an 60\% training and 40\% test split, with training sizes specified in Table~\ref{tab:r2_results_ranked_ties_compact}. Hyperparameters are tuned on the full training set following the procedure in Section~\ref{Sec5.1b}. To ensure reliability, each experiment is repeated 10 times independently across the four datasets. The results are summarized in Table~\ref{tab:r2_results_ranked_ties_compact}.

\subsection{Results}
First, to demonstrate o1Neuro's practical effectiveness, the left panel of Table~\ref{tab:r2_results_ranked_ties_compact} without transforming features shows that its average rank across the four datasets is $1.5 = (1 + 1 + 2 + 2) / 4$, the best among the four models (Random Forests: $1.75$, XGBoost: $2.25$, TabNet: $3.75$). In the elevators dataset, Random Forests lag behind o1Neuro and XGBoost by a significant margin ($\ge 10\%$ in R$^2$), while in abalone, XGBoost performs noticeably worse than both o1Neuro and Random Forests. TabNet performs reasonably well on abalone but remains inaccurate and unstable in other datasets.

Next, focusing on the right panel of Table~\ref{tab:r2_results_ranked_ties_compact}, where we introduce complex interaction components by transforming the original input features, o1Neuro achieves an even stronger performance. It ranks first overall with an average rank of $1.25$ and demonstrates clear superiority, especially in the elevators and Ailerons datasets, outperforming all other models by substantial margins ($\ge 10\%$ in R$^2$ score).

In summary, the empirical results in this section, together with those in Section~\ref{Sec6b}, provide strong evidence that o1Neuro delivers significant advantages when modeling complex relationships between variables, as supported by Theorem~\ref{lemma1}, while maintaining a reasonable and tunable computational cost.

\section{Reproducibility and Code Availability}\label{Sec8}

The real datasets used in Section~\ref{Sec7b} are publicly available from OpenML~\citep{OpenML2013} and the UCI Machine Learning Repository~\citep{Dua:2017}. The Python implementation of o1Neuro is publicly available at \url{https://github.com/xbb66kw/o1Neuro}.



\bibliography{references}   

\begin{thebibliography}{61}
\providecommand{\natexlab}[1]{#1}
\providecommand{\url}[1]{\texttt{#1}}
\expandafter\ifx\csname urlstyle\endcsname\relax
  \providecommand{\doi}[1]{doi: #1}\else
  \providecommand{\doi}{doi: \begingroup \urlstyle{rm}\Url}\fi

\bibitem[Adcock and Dexter(2021)]{adcock2021gap}
Ben Adcock and Nick Dexter.
\newblock The gap between theory and practice in function approximation with deep neural networks.
\newblock \emph{SIAM Journal on Mathematics of Data Science}, 3\penalty0 (2):\penalty0 624--655, 2021.

\bibitem[Arik and Pfister(2021)]{arik2021tabnet}
Sercan~{\"O} Arik and Tomas Pfister.
\newblock Tabnet: Attentive interpretable tabular learning.
\newblock In \emph{Proceedings of the AAAI conference on artificial intelligence}, volume~35, pages 6679--6687, 2021.

\bibitem[Arora et~al.(2019)Arora, Du, Hu, Li, Salakhutdinov, and Wang]{arora2019exact}
Sanjeev Arora, Simon~S Du, Wei Hu, Zhiyuan Li, Russ~R Salakhutdinov, and Ruosong Wang.
\newblock On exact computation with an infinitely wide neural net.
\newblock \emph{Advances in neural information processing systems}, 32, 2019.

\bibitem[Barron(1992)]{barron1992neural}
Andrew~R Barron.
\newblock Neural net approximation.
\newblock In \emph{Proc. 7th Yale workshop on adaptive and learning systems}, volume~1, pages 69--72, 1992.

\bibitem[Barron(1993)]{barron1993universal}
Andrew~R Barron.
\newblock Universal approximation bounds for superpositions of a sigmoidal function.
\newblock \emph{IEEE Transactions on Information theory}, 39\penalty0 (3):\penalty0 930--945, 1993.

\bibitem[Barron and Luo(2007)]{barron2007adaptive}
Andrew~R Barron and Xi~Luo.
\newblock Adaptive annealing.
\newblock In \emph{Proceedings of the Allerton Conference on Communications, Computation and Control}, pages 665--673, 2007.

\bibitem[Barron et~al.(2008)Barron, Cohen, Dahmen, and DeVore]{barron2008approximation}
Andrew~R. Barron, Albert Cohen, Wolfgang Dahmen, and Ronald~A. DeVore.
\newblock {Approximation and learning by greedy algorithms}.
\newblock \emph{The Annals of Statistics}, 36\penalty0 (1):\penalty0 64 -- 94, 2008.
\newblock \doi{10.1214/009053607000000631}.
\newblock URL \url{https://doi.org/10.1214/009053607000000631}.

\bibitem[Bauer and Kohler(2019)]{bauer2019deep}
Benedikt Bauer and Michael Kohler.
\newblock On deep learning as a remedy for the curse of dimensionality in nonparametric regression.
\newblock \emph{Ann. Statist. 47 (4) 2261 - 2285, August 2019.}, 2019.

\bibitem[Biau and Cadre(2021)]{biau2021optimization}
G{\'e}rard Biau and Beno{\^\i}t Cadre.
\newblock Optimization by gradient boosting.
\newblock In \emph{Advances in Contemporary Statistics and Econometrics: Festschrift in Honor of Christine Thomas-Agnan}, pages 23--44. Springer, 2021.

\bibitem[Bien et~al.(2013)Bien, Taylor, and Tibshirani]{bien2013lasso}
Jacob Bien, Jonathan Taylor, and Robert Tibshirani.
\newblock A lasso for hierarchical interactions.
\newblock \emph{Annals of st atistics}, 41\penalty0 (3):\penalty0 1111, 2013.

\bibitem[Breiman(2001)]{breiman2001random}
Leo Breiman.
\newblock Random forests.
\newblock \emph{Machine learning}, 45:\penalty0 5--32, 2001.

\bibitem[Brock et~al.(2017)Brock, Lim, Ritchie, and Weston]{brock2017freezeout}
Andrew Brock, Theodore Lim, James~M Ritchie, and Nick Weston.
\newblock Freezeout: Accelerate training by progressively freezing layers.
\newblock \emph{arXiv preprint arXiv:1706.04983}, 2017.

\bibitem[Brown et~al.(2020)Brown, Mann, Ryder, Subbiah, Kaplan, Dhariwal, Neelakantan, Shyam, Sastry, Askell, et~al.]{brown2020language}
Tom Brown, Benjamin Mann, Nick Ryder, Melanie Subbiah, Jared~D Kaplan, Prafulla Dhariwal, Arvind Neelakantan, Pranav Shyam, Girish Sastry, Amanda Askell, et~al.
\newblock Language models are few-shot learners.
\newblock \emph{Advances in neural information processing systems}, 33:\penalty0 1877--1901, 2020.

\bibitem[Cao and Gu(2019)]{cao2019generalization}
Yuan Cao and Quanquan Gu.
\newblock Generalization bounds of stochastic gradient descent for wide and deep neural networks.
\newblock \emph{Advances in neural information processing systems}, 32, 2019.

\bibitem[Chen and Guestrin(2016)]{chen2016xgboost}
Tianqi Chen and Carlos Guestrin.
\newblock Xgboost: A scalable tree boosting system.
\newblock In \emph{Proceedings of the 22nd acm sigkdd international conference on knowledge discovery and data mining}, pages 785--794, 2016.

\bibitem[Choromanska et~al.(2015)Choromanska, Henaff, Mathieu, Arous, and LeCun]{choromanska2015loss}
Anna Choromanska, Mikael Henaff, Michael Mathieu, G{\'e}rard~Ben Arous, and Yann LeCun.
\newblock The loss surfaces of multilayer networks.
\newblock In \emph{Artificial intelligence and statistics}, pages 192--204. PMLR, 2015.

\bibitem[Cohn(2013)]{cohn2013measure}
Donald~L Cohn.
\newblock \emph{Measure theory}, volume~2.
\newblock Springer, 2013.

\bibitem[Cox(1984)]{cox1984interaction}
David~R Cox.
\newblock Interaction.
\newblock \emph{International Statistical Review/Revue Internationale de Statistique}, pages 1--24, 1984.

\bibitem[Cybenko(1989)]{cybenko1989approximation}
George Cybenko.
\newblock Approximation by superpositions of a sigmoidal function.
\newblock \emph{Mathematics of control, signals and systems}, 2\penalty0 (4):\penalty0 303--314, 1989.

\bibitem[D{\'e}fossez et~al.(2020)D{\'e}fossez, Bottou, Bach, and Usunier]{defossez2020simple}
Alexandre D{\'e}fossez, L{\'e}on Bottou, Francis Bach, and Nicolas Usunier.
\newblock A simple convergence proof of adam and adagrad.
\newblock \emph{arXiv preprint arXiv:2003.02395}, 2020.

\bibitem[DeVore and Temlyakov(1996)]{devore1996some}
Ronald~A DeVore and Vladimir~N Temlyakov.
\newblock Some remarks on greedy algorithms.
\newblock \emph{Advances in computational Mathematics}, 5\penalty0 (1):\penalty0 173--187, 1996.

\bibitem[Duchi et~al.(2011)Duchi, Hazan, and Singer]{duchi2011adaptive}
John Duchi, Elad Hazan, and Yoram Singer.
\newblock Adaptive subgradient methods for online learning and stochastic optimization.
\newblock \emph{Journal of machine learning research}, 12\penalty0 (7), 2011.

\bibitem[Elman(1990)]{elman1990finding}
Jeffrey~L Elman.
\newblock Finding structure in time.
\newblock \emph{Cognitive science}, 14\penalty0 (2):\penalty0 179--211, 1990.

\bibitem[Fan et~al.(2020)Fan, Ma, and Zhong]{fan2020selective}
Jianqing Fan, Cong Ma, and Yiqiao Zhong.
\newblock A selective overview of deep learning.
\newblock \emph{Statistical science: a review journal of the Institute of Mathematical Statistics}, 36\penalty0 (2):\penalty0 264, 2020.

\bibitem[Farag{\'o} and Lugosi(1993)]{farago1993strong}
Andr{\'a}s Farag{\'o} and G{\'a}bor Lugosi.
\newblock Strong universal consistency of neural network classifiers.
\newblock \emph{IEEE Transactions on Information Theory}, 39\penalty0 (4):\penalty0 1146--1151, 1993.

\bibitem[Fehrman et~al.(2020)Fehrman, Gess, and Jentzen]{fehrman2020convergence}
Benjamin Fehrman, Benjamin Gess, and Arnulf Jentzen.
\newblock Convergence rates for the stochastic gradient descent method for non-convex objective functions.
\newblock \emph{Journal of Machine Learning Research}, 21\penalty0 (136):\penalty0 1--48, 2020.

\bibitem[Friedman(2001)]{friedman2001greedy}
Jerome~H Friedman.
\newblock Greedy function approximation: a gradient boosting machine.
\newblock \emph{Annals of statistics}, pages 1189--1232, 2001.

\bibitem[Gentile and Welper(2024)]{gentile2024approximation}
Russell Gentile and Gerrit Welper.
\newblock Approximation results for gradient flow trained shallow neural networks in 1d.
\newblock \emph{Constructive Approximation}, 60\penalty0 (3):\penalty0 547--594, 2024.

\bibitem[Grinsztajn et~al.(2022)Grinsztajn, Oyallon, and Varoquaux]{grinsztajn2022tree}
L{\'e}o Grinsztajn, Edouard Oyallon, and Ga{\"e}l Varoquaux.
\newblock Why do tree-based models still outperform deep learning on typical tabular data?
\newblock \emph{Advances in neural information processing systems}, 35:\penalty0 507--520, 2022.

\bibitem[Grohs and Voigtlaender(2024)]{grohs2024proof}
Philipp Grohs and Felix Voigtlaender.
\newblock Proof of the theory-to-practice gap in deep learning via sampling complexity bounds for neural network approximation spaces.
\newblock \emph{Foundations of Computational Mathematics}, 24\penalty0 (4):\penalty0 1085--1143, 2024.

\bibitem[Gy{\"o}rfi et~al.(2002)Gy{\"o}rfi, Kohler, Krzy{\.z}ak, and Walk]{gyorfi2002distribution}
L{\'a}szl{\'o} Gy{\"o}rfi, Michael Kohler, Adam Krzy{\.z}ak, and Harro Walk.
\newblock \emph{A distribution-free theory of nonparametric regression}.
\newblock Springer, 2002.

\bibitem[Herrmann et~al.(2022)Herrmann, Opschoor, and Schwab]{herrmann2022constructive}
Lukas Herrmann, Joost~AA Opschoor, and Christoph Schwab.
\newblock Constructive deep relu neural network approximation.
\newblock \emph{Journal of Scientific Computing}, 90\penalty0 (2):\penalty0 75, 2022.

\bibitem[Hinton(2012)]{hinton2012neural}
Geoffrey Hinton.
\newblock Neural networks for machine learning — lecture 6a: Overview of mini-batch gradient descent, 2012.
\newblock URL \url{https://www.coursera.org/learn/neural-networks-machine-learning}.
\newblock Coursera Course.

\bibitem[Hornik et~al.(1989)Hornik, Stinchcombe, and White]{hornik1989multilayer}
Kurt Hornik, Maxwell Stinchcombe, and Halbert White.
\newblock Multilayer feedforward networks are universal approximators.
\newblock \emph{Neural networks}, 2\penalty0 (5):\penalty0 359--366, 1989.

\bibitem[Jacot et~al.(2018)Jacot, Gabriel, and Hongler]{jacot2018neural}
Arthur Jacot, Franck Gabriel, and Cl{\'e}ment Hongler.
\newblock Neural tangent kernel: Convergence and generalization in neural networks.
\newblock \emph{Advances in neural information processing systems}, 31, 2018.

\bibitem[Jentzen and Riekert(2022)]{jentzen2022proof}
Arnulf Jentzen and Adrian Riekert.
\newblock A proof of convergence for the gradient descent optimization method with random initializations in the training of neural networks with relu activation for piecewise linear target functions.
\newblock \emph{Journal of Machine Learning Research}, 23\penalty0 (260):\penalty0 1--50, 2022.

\bibitem[Jentzen and Riekert(2024)]{jentzen2024non}
Arnulf Jentzen and Adrian Riekert.
\newblock Non-convergence to global minimizers for adam and stochastic gradient descent optimization and constructions of local minimizers in the training of artificial neural networks.
\newblock \emph{arXiv preprint arXiv:2402.05155}, 2024.

\bibitem[Jiao et~al.(2023)Jiao, Shen, Lin, and Huang]{jiao2023deep}
Yuling Jiao, Guohao Shen, Yuanyuan Lin, and Jian Huang.
\newblock Deep nonparametric regression on approximate manifolds: Nonasymptotic error bounds with polynomial prefactors.
\newblock \emph{The Annals of Statistics}, 51\penalty0 (2):\penalty0 691--716, 2023.

\bibitem[Jones(1992)]{jones1992simple}
Lee~K Jones.
\newblock A simple lemma on greedy approximation in hilbert space and convergence rates for projection pursuit regression and neural network training.
\newblock \emph{The annals of Statistics}, pages 608--613, 1992.

\bibitem[Jumper et~al.(2021)Jumper, Evans, Pritzel, Green, Figurnov, Ronneberger, Tunyasuvunakool, Bates, {\v{Z}}{\'\i}dek, Potapenko, et~al.]{jumper2021highly}
John Jumper, Richard Evans, Alexander Pritzel, Tim Green, Michael Figurnov, Olaf Ronneberger, Kathryn Tunyasuvunakool, Russ Bates, Augustin {\v{Z}}{\'\i}dek, Anna Potapenko, et~al.
\newblock Highly accurate protein structure prediction with alphafold.
\newblock \emph{nature}, 596\penalty0 (7873):\penalty0 583--589, 2021.

\bibitem[Kingma and Ba(2015)]{kingma2015adam}
Diederik~P. Kingma and Jimmy Ba.
\newblock Adam: A method for stochastic optimization.
\newblock In \emph{3rd International Conference on Learning Representations (ICLR)}, 2015.
\newblock URL \url{https://arxiv.org/abs/1412.6980}.

\bibitem[Krizhevsky et~al.(2012)Krizhevsky, Sutskever, and Hinton]{krizhevsky2012imagenet}
Alex Krizhevsky, Ilya Sutskever, and Geoffrey~E Hinton.
\newblock Imagenet classification with deep convolutional neural networks.
\newblock \emph{Advances in neural information processing systems}, 25, 2012.

\bibitem[LeCun et~al.(2015)LeCun, Bengio, and Hinton]{lecun2015deep}
Yann LeCun, Yoshua Bengio, and Geoffrey Hinton.
\newblock Deep learning.
\newblock \emph{nature}, 521\penalty0 (7553):\penalty0 436--444, 2015.

\bibitem[Lee et~al.(1996)Lee, Bartlett, and Williamson]{lee1996efficient}
Wee~Sun Lee, Peter~L Bartlett, and Robert~C Williamson.
\newblock Efficient agnostic learning of neural networks with bounded fan-in.
\newblock \emph{IEEE Transactions on Information Theory}, 42\penalty0 (6):\penalty0 2118--2132, 1996.

\bibitem[Li et~al.(2023)Li, Rakhlin, and Jadbabaie]{li2023convergence}
Haochuan Li, Alexander Rakhlin, and Ali Jadbabaie.
\newblock Convergence of adam under relaxed assumptions.
\newblock \emph{Advances in Neural Information Processing Systems}, 36:\penalty0 52166--52196, 2023.

\bibitem[Malladi et~al.(2023)Malladi, Wettig, Yu, Chen, and Arora]{malladi2023kernel}
Sadhika Malladi, Alexander Wettig, Dingli Yu, Danqi Chen, and Sanjeev Arora.
\newblock A kernel-based view of language model fine-tuning.
\newblock In \emph{International Conference on Machine Learning}, pages 23610--23641. PMLR, 2023.

\bibitem[Markelle~Kelly(2017)]{Dua:2017}
Kolby~Nottingham Markelle~Kelly, Rachel~Longjohn.
\newblock The uci machine learning repository, 2017.
\newblock URL \url{http://archive.ics.uci.edu/ml}.

\bibitem[McElfresh et~al.(2023)McElfresh, Khandagale, Valverde, Prasad~C, Ramakrishnan, Goldblum, and White]{mcelfresh2023neural}
Duncan McElfresh, Sujay Khandagale, Jonathan Valverde, Vishak Prasad~C, Ganesh Ramakrishnan, Micah Goldblum, and Colin White.
\newblock When do neural nets outperform boosted trees on tabular data?
\newblock \emph{Advances in Neural Information Processing Systems}, 36:\penalty0 76336--76369, 2023.

\bibitem[Mei et~al.(2018)Mei, Montanari, and Nguyen]{mei2018mean}
Song Mei, Andrea Montanari, and Phan-Minh Nguyen.
\newblock A mean field view of the landscape of two-layer neural networks.
\newblock \emph{Proceedings of the National Academy of Sciences}, 115\penalty0 (33):\penalty0 E7665--E7671, 2018.

\bibitem[Mei et~al.(2019)Mei, Misiakiewicz, and Montanari]{mei2019mean}
Song Mei, Theodor Misiakiewicz, and Andrea Montanari.
\newblock Mean-field theory of two-layers neural networks: dimension-free bounds and kernel limit.
\newblock In \emph{Conference on learning theory}, pages 2388--2464. PMLR, 2019.

\bibitem[Schmidt-Hieber(2020)]{schmidt2020nonparametric}
Johannes Schmidt-Hieber.
\newblock Nonparametric regression using deep neural networks with relu activation function.
\newblock \emph{The Annals of Statistics}, 2020.

\bibitem[Shwartz-Ziv and Armon(2022)]{shwartz2022tabular}
Ravid Shwartz-Ziv and Amitai Armon.
\newblock Tabular data: Deep learning is not all you need.
\newblock \emph{Information Fusion}, 81:\penalty0 84--90, 2022.

\bibitem[Siegel and Xu(2022)]{siegel2022optimal}
Jonathan~W Siegel and Jinchao Xu.
\newblock Optimal convergence rates for the orthogonal greedy algorithm.
\newblock \emph{IEEE Transactions on Information Theory}, 68\penalty0 (5):\penalty0 3354--3361, 2022.

\bibitem[Silver et~al.(2016)Silver, Huang, Maddison, Guez, Sifre, Van Den~Driessche, Schrittwieser, Antonoglou, Panneershelvam, Lanctot, et~al.]{silver2016mastering}
David Silver, Aja Huang, Chris~J Maddison, Arthur Guez, Laurent Sifre, George Van Den~Driessche, Julian Schrittwieser, Ioannis Antonoglou, Veda Panneershelvam, Marc Lanctot, et~al.
\newblock Mastering the game of go with deep neural networks and tree search.
\newblock \emph{nature}, 529\penalty0 (7587):\penalty0 484--489, 2016.

\bibitem[Stein and Shakarchi(2009)]{stein2009real}
Elias~M Stein and Rami Shakarchi.
\newblock \emph{Real analysis: measure theory, integration, and Hilbert spaces}.
\newblock Princeton University Press, 2009.

\bibitem[Telgarsky(2016)]{telgarsky2016benefits}
Matus Telgarsky.
\newblock Benefits of depth in neural networks.
\newblock In \emph{Conference on learning theory}, pages 1517--1539. PMLR, 2016.

\bibitem[Temlyakov(2000)]{temlyakov2000weak}
Vladimir~N Temlyakov.
\newblock Weak greedy algorithms.
\newblock \emph{Advances in Computational Mathematics}, 12\penalty0 (2):\penalty0 213--227, 2000.

\bibitem[Vanschoren et~al.(2013)Vanschoren, van Rijn, Bischl, and Torgo]{OpenML2013}
Joaquin Vanschoren, Jan~N. van Rijn, Bernd Bischl, and Luis Torgo.
\newblock Openml: Networked science in machine learning.
\newblock \emph{SIGKDD Explorations}, 15\penalty0 (2):\penalty0 49--60, 2013.
\newblock \doi{10.1145/2641190.2641198}.
\newblock URL \url{http://doi.acm.org/10.1145/2641190.2641198}.

\bibitem[Xu and Zhu(2024)]{xu2024overparametrized}
Jiaming Xu and Hanjing Zhu.
\newblock Overparametrized multi-layer neural networks: Uniform concentration of neural tangent kernel and convergence of stochastic gradient descent.
\newblock \emph{Journal of Machine Learning Research}, 25\penalty0 (94):\penalty0 1--83, 2024.

\bibitem[Yarotsky(2018)]{yarotsky2018optimal}
Dmitry Yarotsky.
\newblock Optimal approximation of continuous functions by very deep relu networks.
\newblock In \emph{Conference on learning theory}, pages 639--649. PMLR, 2018.

\bibitem[Zhang et~al.(2022)Zhang, Chen, Shi, Sun, and Luo]{zhang2022adam}
Yushun Zhang, Congliang Chen, Naichen Shi, Ruoyu Sun, and Zhi-Quan Luo.
\newblock Adam can converge without any modification on update rules.
\newblock \emph{Advances in neural information processing systems}, 35:\penalty0 28386--28399, 2022.

\end{thebibliography}

\newpage

\appendix

	\setcounter{page}{1}
	\setcounter{section}{0}

	\begin{center}{\bf \Large Supplementary Material to ``Constructive Universal Approximation and Sure Convergence for Multi-Layer Neural Networks''}
		
		\bigskip
		
		Chien-Ming Chi
	\end{center}
	
\noindent The Supplementary Material provides additional content to support the main text. Section~\ref{SecA} presents supplementary details, including hyperparameter tuning procedures and search spaces. Section~\ref{SecB} contains detailed proofs of the main theorems and Example~\ref{example1}, while Section~\ref{SecC} presents technical lemmas along with their proofs.

	\renewcommand{\theequation}{A.\arabic{equation}}
	\renewcommand{\thesubsection}{A.\arabic{subsection}}
	\setcounter{equation}{0}

    \section{Hyperparameter Spaces}\label{SecA}

We use the Python package \texttt{hyperopt} from \url{https://hyperopt.github.io/hyperopt/} for tuning all predictive models. The hyperparameter spaces of o1Neuro, TabNet, XGBoost, Random Forests are respectively given in Table~\ref{tab:validation_space_o1neuro} and Tables~\ref{tab:tabnet_space}--\ref{tab:validation_space_rf}. The hyperparameter search space for TabNet~\citep{arik2021tabnet} is slightly restricted to reduce runtime.

\begin{table}[htbp]
\centering
\caption{
Hyperparameter search space for TabNet. 
$n_d$: dimension of decision prediction; 
$n_a$: dimension of attention prediction; 
$n_{\text{steps}}$: number of sequential steps in the feature transformer; 
$\gamma$: relaxation factor controlling attention sparsity; 
$\lambda_{\text{sparse}}$: sparsity regularization coefficient; 
$\text{learning\_rate}$: step size for gradient updates; 
$\text{batch\_size}$: number of samples per batch; 
$\text{virtual\_batch\_size}$: sub-batch size for virtual batch normalization.
}
\label{tab:tabnet_space}
\begin{tabular}{ll}
\hline
Parameter name & Search Space \\
\hline
$n_d$                       & $\{8,\,16,\,24\}$ \\
$n_a$                       & $\{8,\,16,\,24\}$ \\
$n_{\text{steps}}$          & $\{3,\,4,\,5\}$ \\
$\gamma$                    & $\text{Uniform}(1.0,\,1.8)$ \\
$\lambda_{\text{sparse}}$   & $\text{LogUniform}(10^{-5},\,10^{-2})$ \\
$\text{learning\_rate}$     & $\text{LogUniform}(10^{-3},\,0.02)$ \\
$\text{batch\_size}$        & $\{ 64,\, 128,\, 256,\,512\}$ \\ 
$\text{virtual\_batch\_size}$ & $\{64,\,128\}$ \\
\hline
\end{tabular}
\end{table}

\begin{table}[htbp]
\centering
\caption{
Hyperparameter search space for XGBoost. 
$n\_estimators$: number of trees; 
$\gamma$: minimum loss reduction to split a leaf; 
$\text{reg\_alpha}$: L1 regularization; 
$\text{reg\_lambda}$: L2 regularization; 
$\text{learning\_rate}$: step size shrinkage; 
$\text{subsample}$: fraction of samples per tree; 
$\text{colsample\_bytree}$: fraction of features per tree; 
$\text{colsample\_bylevel}$: fraction of features per split; 
$\text{min\_child\_weight}$: minimum sum of Hessian in a child; 
$\text{max\_depth}$: maximum tree depth.
}

\label{tab:validation_space_xgb}
\begin{tabular}{ll}
\hline
Parameter name & Search Space \\
\hline
$n\_\text{estimators}$ & 1000 \\
$\gamma$ (min\_split\_loss) & $\exp(Z), Z \sim \text{Uniform}[-8\log 10, \log 7]$ \\
$\text{reg\_alpha}$ & $\exp(Z), Z \sim \text{Uniform}[-8\log 10, \log 100]$ \\
$\text{reg\_lambda}$ & $\exp(Z), Z \sim \text{Uniform}[\log 0.8, \log 4]$ \\
$\text{learning\_rate}$ & $\exp(Z), Z \sim \text{Uniform}[-5\log 10, \log 0.7]$ \\
$\text{subsample}$ & Uniform $[0.5, 1]$ \\
$\text{colsample\_bytree}$ & Uniform $[0.5, 1]$ \\
$\text{colsample\_bylevel}$ & Uniform $[0.5, 1]$ \\
$\text{min\_child\_weight}$ & Uniform $\{0, \dots, 20\}$ \\
$\text{max\_depth}$ & Uniform $\{2, \dots, 15\}$ \\
\hline
\end{tabular}
\end{table}

\begin{table}[htbp]
\centering
\caption{
Hyperparameter search space for Random Forests. 
$n\_estimators$: number of trees; 
$\text{max\_depth}$: maximum depth of each tree; 
$\text{min\_samples\_split}$: minimum samples required to split a node; 
$\text{min\_samples\_leaf}$: minimum samples required at a leaf node; 
$\text{min\_impurity\_decrease}$: minimum decrease in impurity to split a node; 
$\text{criterion}$: function to measure split quality.
}
\label{tab:validation_space_rf}
\begin{tabular}{ll}
\hline
Parameter name & Search Space \\
\hline
$n\_\text{estimators}$ & 100 \\
$\gamma$ (column subsampling rate)   & Uniform $[0, 1]$ \\
$\text{min\_samples\_split}$ & $\{1, \dots, 20\}$ \\
$\text{min\_samples\_leaf}$ & $\{2, \dots, 20\}$ \\
$\text{min\_impurity\_decrease}$ & $\{0, 0.01, 0.02, 0.05\}$ \\
$\text{max\_depth}$ & $\{5, 10, 20, 50, \infty\}$ \\
$\text{criterion}$ & $\{\text{squared\_error}, \text{absolute\_error}\}$ \\
\hline
\end{tabular}
\end{table}

    \renewcommand{\theequation}{B.\arabic{equation}}
	\renewcommand{\thesubsection}{B.\arabic{subsection}}
	\setcounter{equation}{0}
\section{Proofs of Main Results}\label{SecB}

\subsection{Proof of Theorem~\ref{lemma1}}

We begin with proving the first assertion of Theorem~\ref{lemma1}. Owing to the sparse structure of the o1Neuro network $\norm{\vv{w}_{l, h}}_{0}\le 2$ and the set-up $p_{l} \ge 2^{L - l}\, M$ for $l \in \{1, \dots, L - 1\}$, the function class based on 
\[
(f_{L,1}(\vv{x}), \dots, f_{L,M}(\vv{x}))
\]
is equivalent to the class based on 
\[
(g_1(\vv{x}), \dots, g_M(\vv{x})), \quad g_h \in \widetilde{\mathcal{G}}.
\]
As a result, together with the optimization procedure described in Section~\ref{Sec2.1.1}, we conclude that the first assertion of Theorem~\ref{lemma1} holds. Particularly, the result holds with $t = 1$. Note that when $t <1$, it is also referred to as the weak (boosted) greedy algorithm~\citep{temlyakov2000weak}.

To proceed, we need the following Lemma~\ref{lemma3}, whose proof is given in Section~\ref{proof.lemma3}. Recall that $\mathcal{L}(k)$ has been defined in Condition~\ref{condi.inde}.
\begin{lemma}\label{lemma3}
    If $L \ge 2 + \lceil \log_2 k \rceil$, then for each integer $R_0 > 0$, it holds that $\mathcal{L}(k, R_0) \subset \overline{\mathcal{D}},$
where
\begin{equation*}
    \begin{split}
\mathcal{L}(k, R_0) & = \left\{ \sum_{r=1}^{R_0} f_r : f_r \in \mathcal{L}(k) \right\},        \\
\overline{\mathcal{D}} & \coloneqq \left\{ f : \mathbb{R}^p \to \mathbb{R} \mid \lim_{i \to \infty} \mathbb{E}[f_i(\boldsymbol{X}) - f(\boldsymbol{X})]^2 = 0 \text{ for some } f_i \in \mathcal{D} \right\},\\
\mathcal{D} &= \left\{ \sum_{s=1}^{s_0} \sum_{l=0}^1 z_{l,s} \, \boldsymbol{1}\{f_s(\vv{x}) = l\} \mid z_{l,s} \in \mathbb{R}, f_s \in \widetilde{\mathcal{G}}, s_0 \in \mathbb{N} \right\},
    \end{split}
\end{equation*}
with $\mathbb{N}$ denoting the set of positive integers.
\end{lemma}


We now prove the second assertion of Theorem~\ref{lemma1}, assuming that $f_{L,1}, \dots, f_{L,M}$ satisfy \eqref{eq.1} for some $t \in (0,1]$. The proof of Theorem \ref{lemma1} closely follows that of Theorem 1 in \citep{temlyakov2000weak}, but with a major distinction stemming from its application to boosting prediction. We provide a self-contained proof below, employing only auxiliary lemmas from \citep{temlyakov2000weak}.
    
	We recursively define random variables $\{\widetilde{A}_{s}\}_{s\ge 0}$ with $\widetilde{A}_{0} = \mathbb{E}(Y\mid \boldsymbol{X})$ such that for $s  >0$,
		\begin{equation}
			\begin{split}\label{theorem.appr.3}
				\widetilde{A}_{s} & \coloneqq \mathbb{E}(Y\mid \boldsymbol{X}) - \gamma\sum_{q=1}^{s}  \sum_{l=0}^{1} \frac{\boldsymbol{1}\{ f_{L, q}(\boldsymbol{X}) = l\} }{\sqrt{\mathbb{P}(f_{L, q}(\boldsymbol{X}) =l) }} \times \mathbb{E}( \frac{\boldsymbol{1}\{ f_{L, q}(\boldsymbol{X}) = l\} }{\sqrt{\mathbb{P}(f_{L, q}(\boldsymbol{X}) =l) }} \widetilde{A}_{q-1}),            
			\end{split}
		\end{equation}
        where $f_{L, s}$'s are given by Theorem~\ref{lemma1}.
        The definitions of $\widetilde{A}_{s}$ and $\widetilde{R}_{s}$ coincides for $s>0$, which is justified as follows. By \eqref{theorem.appr.3} and the fact that for each $g\in\widetilde{\mathcal{G}}$, it holds that 
        \begin{equation*}
            \begin{split}
                \mathbb{E}( \boldsymbol{1}\{ g(\boldsymbol{X}) = l\} \times \widetilde{A}_{0}) & = \mathbb{E}\{ \boldsymbol{1}\{ g(\boldsymbol{X}) = l\} \times [\widetilde{R}_{0} - (Y - \mathbb{E}(Y\mid \boldsymbol{X}))] \} \\
                & = \mathbb{E}\{ \boldsymbol{1}\{ g(\boldsymbol{X}) = l\} \times \widetilde{R}_{0}  \}
            \end{split}
        \end{equation*}
due to the definition $\widetilde{R}_0 = Y$ in Section~\ref{Sec2.1.1} and the law of total expectation, we conclude for each $s> 0$ that 
\begin{equation}
    \label{B0}
    \widetilde{R}_s = \widetilde{A}_s \text{ almost surely.}
\end{equation}

		On the other hand, a direct calculation shows that for $s>q\ge 0$,
		\begin{equation}
			\begin{split}\label{theorem.appr.4}
				\mathbb{E}(\widetilde{A}_{s})^2 & = \mathbb{E}(\widetilde{A}_{s-1})^2 + \gamma^2 \times \mathcal{P}_{s}(f_{L, s}) - 2 \gamma \times \mathcal{P}_{s}(f_{L, s}) \\
				& = \mathbb{E}(\widetilde{A}_{q})^2  - \sum_{l=q+1}^{s}\gamma(2-\gamma)\mathcal{P}_{l}(f_{L, l})\\
				& = \mathbb{E}[\mathbb{E}(Y\mid \boldsymbol{X})]^2  - \gamma(2-\gamma)\sum_{l=1}^{s}\mathcal{P}_{l}(f_{L, l}),
			\end{split}
		\end{equation}
        where for each $g\in \widetilde{\mathcal{G}}$ and $s> 0$, we define
$$\mathcal{P}_s (g) \coloneqq \sum_{l=0}^{1} \left[\mathbb{E}( \frac{\boldsymbol{1}\{ g(\boldsymbol{X}) = l\} }{\sqrt{\mathbb{P}(g(\boldsymbol{X}) =l) }} \widetilde{A}_{s-1}) \right]^2.$$

		With \eqref{theorem.appr.4}, we deduce that 
		\begin{equation}\label{theorem.appr.7}
			\sum_{q=1}^{\infty}\mathcal{P}_{q}(f_{L, q}) \le \frac{\mathbb{E}[\mathbb{E}(Y\mid \boldsymbol{X})]^2}{\gamma(2-\gamma)} < \infty.
		\end{equation}
        Additionally, a direct calculation shows that for every $s>q$,
		\begin{equation}
				\begin{split}\label{theorem.appr.5}
					\mathbb{E}(\widetilde{A}_{s} - \widetilde{A}_{q})^2  
					& = \mathbb{E}(\widetilde{A}_{q})^2  - \mathbb{E}(\widetilde{A}_{s})^2 - 2\mathbb{E}[(\widetilde{A}_{q} - \widetilde{A}_{s}) \widetilde{A}_{s} ].
				\end{split}
		\end{equation}

Using \eqref{theorem.appr.7}–\eqref{theorem.appr.5} and Lemma 2.4 of \citep{temlyakov2000weak}, it suffices to show that the absolute value of the second term in \eqref{theorem.appr.5} tends to zero as \eqref{convergence.sequence} below, in order to conclude that $\widetilde{A}_s$ converges in the $L_2$-norm (as in \eqref{theorem.appr.10} below).

		For each $i>0$ and each $l\in \{0, 1\}$, denote $a_i(l) = \left|\mathbb{E}( \frac{\boldsymbol{1}\{f_{L, i}(\boldsymbol{X}) = l\}}{ \sqrt{\mathbb{P}(f_{L, i}(\boldsymbol{X}) = l) } } \widetilde{A}_{i-1} )\right|$. In light of \eqref{theorem.appr.3}--\eqref{B0} and \eqref{eq.1}, we deduce that for every $s> q$,
		\begin{equation}
			\begin{split}\label{theorem.appr.6}
				& |\mathbb{E}[(\widetilde{A}_{q} - \widetilde{A}_{s}) \widetilde{A}_{s} ] | \\
				& \le \gamma\sum_{i=q+1}^{s}  \sum_{l=0}^{1} \frac{ |\mathbb{E}(\widetilde{A}_{s}\times \boldsymbol{1}\{f_{L, i}(\boldsymbol{X}) = l\})| }{\sqrt{\mathbb{P}(f_{L, i}(\boldsymbol{X}) = l) }} \times a_{i}(l) \\
				& \le \gamma\sum_{i=q+1}^{s} \sqrt{ \mathcal{P}_{s+1}(f_{L, s} ) } \sum_{l=0}^{1}   a_{i}(l) \\
				& \le \gamma\sum_{i=q+1}^{s} \sqrt{  \mathcal{P}_{s+1}(f_{L, s+1}) t^{-1}} \sum_{l=0}^{1}   a_{i}(l)\\
				& \le \gamma \sqrt{  \mathcal{P}_{s+1}(f_{L, s+1}) t^{-1}}\sum_{i=q+1}^{s} \sqrt{ 2 \mathcal{P}_{i}(f_{L, i})}.
			\end{split}
		\end{equation}
		The second inequality holds because
		$$\frac{ |\mathbb{E}(\widetilde{A}_{s}\times \boldsymbol{1}\{g(\boldsymbol{X}) = l\})| }{\sqrt{\mathbb{P}(g(\boldsymbol{X}) = l) }} \le \sqrt{\sum_{l=0}^{1} [\frac{ |\mathbb{E}(\widetilde{A}_{s}\times \boldsymbol{1}\{g(\boldsymbol{X}) = l\})| }{\sqrt{\mathbb{P}(g(\boldsymbol{X}) = l) }}]^2 } = \sqrt{ \mathcal{P}_{s+1}(g) }$$ 
		for each $g \in \widetilde{\mathcal{G}}$. The third inequality follows from that the definition of $f_{L, s+1}$ , \eqref{eq.1}, and \eqref{B0}. The fourth inequlaity results from the definition of $a_{i}(l)$ and that
		\begin{equation*}
			\begin{split}
				\left(\sum_{l=0}^{1}   a_{i}(l) \right)^2 \le 2\sum_{l=0}^{1}  \left( a_{i}(l) \right)^2 =  2\mathcal{P}_{i}(f_{L, i}).
			\end{split}    
		\end{equation*}

		By \eqref{theorem.appr.7}, Lemma 2.3 of \citep{temlyakov2000weak}, and \eqref{theorem.appr.6}, we conclude that  
        \begin{equation}
            \label{convergence.sequence}
            \lim_{s\rightarrow\infty}\max_{q< s}\left|\mathbb{E}[(\widetilde{A}_q) - \widetilde{A}_s) \widetilde{A}_s] \right| = 0.
        \end{equation}        
        Furthermore, from \eqref{theorem.appr.4}--\eqref{theorem.appr.7} and the Monotone Convergence Theorem for sequences of real numbers, it follows that \( \mathbb{E}(\widetilde{A}_s)^2 \) converges. Combining the result of \eqref{convergence.sequence}, the convergence of $\mathbb{E}(\widetilde{A}_s)^2$, \eqref{theorem.appr.5}, and Lemma 2.4 from \citep{temlyakov2000weak}, we conclude that $\widetilde{A}_s$ converges in the $L_2$-norm.

		Next, we prove by contradiction that $\widetilde{A}_{s}$ converges to the zero function with $\lim_{s\rightarrow\infty}\mathbb{E}(\widetilde{A}_{s})^2 = 0$. By Condition~\ref{condi.inde}, Lemma~\ref{lemma3}, and the definition of $\overline{\mathcal{D}}$, both $m^{\star}$ and each $m_s$, where $m^{\star}(\boldsymbol{X}) = \mathbb{E}(Y \mid \boldsymbol{X})$ and $m_s(\boldsymbol{X}) = \widetilde{A}_s$ almost surely, lie in $\overline{\mathcal{D}}$ by construction. In light of the definition of $\overline{\mathcal{D}}$ and that $\widetilde{A}_s$ converges in the $L_2$-norm, we conclude that
		\begin{equation}
			\label{theorem.appr.10}
			\lim_{s\rightarrow\infty }\mathbb{E}[\widetilde{A}_{s} - \nu(\boldsymbol{X})]^2 =0
		\end{equation}
		for some measurable function $\nu:\mathbb{R}^{p}\longmapsto\mathbb{R}$ in $\overline{\mathcal{D}}$. In what follows, we consider the nontrivial case where \(\mathbb{E}[\nu(\boldsymbol{X})]^2 > 0\).

		For some $\delta >0$, some $g\in\widetilde{\mathcal{G}}$, and some $l\in \{0, 1\}$, it holds that 
		\begin{equation}
			\label{theorem.appr.9}
			| \mathbb{E}( \nu(\boldsymbol{X})\times \boldsymbol{1}\{g(\boldsymbol{X}) = l\} ) | \ge 2\delta.
		\end{equation}
		For if not, then $\mathbb{E}(\nu(\boldsymbol{X})\times \boldsymbol{1}\{g(\boldsymbol{X}) = l\} ) = 0$ for every $g\in \widetilde{\mathcal{G}}$ and each $l\in \{0, 1\}$, implying that $\mathbb{E}[f_{m}(\boldsymbol{X}) - \nu(\boldsymbol{X})]^2 = \mathbb{E}[f_m(\boldsymbol{X})]^2  +\mathbb{E}[\nu(\boldsymbol{X})]^2 \ge \mathbb{E}[\nu(\boldsymbol{X})]^2  >0$ for every $f_m \in \mathcal{D}$. Such a result contradicts the fact that $\nu(\boldsymbol{X}) \in \overline{\mathcal{D}}$, i.e., there exist $f_m \in \mathcal{D}$ with $\mathbb{E}[(f_m(\boldsymbol{X}) - \nu(\boldsymbol{X}))^2] \to 0$.

		By \eqref{theorem.appr.10}--\eqref{theorem.appr.9} and the definitions of $\widetilde{\mathcal{G}}$, there exists $N_0>0$, $l\in \{0, 1\}$, and $g \in \widetilde{\mathcal{G}}$ such that $| \mathbb{E}( \boldsymbol{1}\{g(\boldsymbol{X}) = l\} \widetilde{A}_q) | \ge \delta$ for all $q > N_0$, if \(\mathbb{E}[\nu(\boldsymbol{X})]^2 > 0\). By this result and that $\mathbb{P}(g(\boldsymbol{X}) = l) \le 1$, we have that for all $q > N_0$, 
		\begin{equation}
			\label{theorem.appr.15}
			\sup_{g\in \widetilde{\mathcal{G}}}\mathcal{P}_{q+1}(g) \ge \delta^2.
		\end{equation}

		By the definition of $f_{L, s}$'s, \eqref{theorem.appr.4}, and \eqref{theorem.appr.15},  we derive that for each $s > q > N_{0}$,
		\begin{equation}
			\begin{split}\label{theorem.appr.2}
				\mathbb{E}(\widetilde{A}_{s})^2 & \le  \mathbb{E}(\widetilde{A}_{q})^2  -  \gamma(2-\gamma) \delta^2 \sum_{i>q}^s t^{-2}.
			\end{split}
		\end{equation}
		This result contradicts the non-negativity of \(\mathbb{E}(\widetilde{A}_{s})^2 \ge 0\). Hence, we conclude \(\mathbb{E}[\nu(\boldsymbol{X})]^2 = 0\) and the desired result  that $\lim_{s\rightarrow\infty}\mathbb{E}(\widetilde{A}_{s})^2 = 0$. 

	By this result, \eqref{theorem.appr.3}, \eqref{B0}, and the definition of $\widetilde{m}$, we have completed the proof of Theorem~\ref{lemma1}.

\subsection{Proof of Theorem~\ref{lemma2}}\label{proof.theorem2}

Suppose the previous $h-1$ output neurons $f_{L, 1}, \dots, f_{L, h-1}$ have been optimized, in the sense of \eqref{eq.2}, implying that for every $j\in \{1, \dots, h-1\}$ and each $g\in \widehat{\mathcal{G}}$,
\begin{equation}
    \begin{split}   \label{prob.2}  
    & (1 + \epsilon_0)\times\widehat{W}_j(f_{L, j}) \\
     & = (1 + \epsilon_0)\sum_{l=0}^1 \frac{ \left[ \frac{1}{n}\sum_{i=1}^{n}\widehat{R}_{i, j-1}  \boldsymbol{1}\{f_{L, j} (\boldsymbol{X}_{i}) = l\} \right]^2}{ \frac{1}{n}\sum_{i=1}^{n}\boldsymbol{1}\{f_{L,j}(\boldsymbol{X}_{i}) = l\}} \\
     & \ge \sum_{l=0}^1 \frac{ \left[ \frac{1}{n}\sum_{i=1}^{n}\widehat{R}_{i, j-1}  \boldsymbol{1}\{g (\boldsymbol{X}_{i}) = l\} \right]^2}{\frac{1}{n}\sum_{i=1}^{n}\boldsymbol{1}\{g (\boldsymbol{X}_{i}) = l\}}.
     \end{split}
\end{equation}
By this and the definition of the sample-level optimization procedure in Section~\ref{Sec2.2.1}, neurons in the subnetworks of $f_{L,1}, \dots, f_{L,h-1}$ are no longer updated; their weights and biases remain fixed in the subsequent update rounds.

From now on, we suppose that subnetworks associated with $f_{L,1}, \dots, f_{L,h-1}$ are optimized at some round $b_1$. Additionally, define for $h\in \{1, \dots, M\}$ that
$$\widehat{\mathcal{G}}_h = \left\{ \text{all } f_{L,h} \text{ satisfying } \eqref{restriction.2} \text{ with subnetworks for } f_{L,1}, \dots, f_{L,h-1} \text{ fixed} \right\}.$$

By Lemma~\ref{lemma10} in Section~\ref{proof.lemma10}, the probability that a randomly sampled $g \in \widehat{\mathcal{G}}_h$ satisfies
\begin{equation}
    \label{prob.1}
    \begin{split}        
    g & = \argmax_{f\in \widehat{\mathcal{G}}_h }\sum_{l=0}^1 \frac{\left[ \frac{1}{n}\sum_{i=1}^{n}\widehat{R}_{i, h-1} \times \boldsymbol{1}\{f (\boldsymbol{X}_{i}) = l\} \right]^2}{\frac{1}{n}\sum_{i=1}^{n}\boldsymbol{1}\{f (\boldsymbol{X}_{i}) = l\}} \\
     & = \argmax_{f\in \widehat{\mathcal{G}} }\sum_{l=0}^1 \frac{\left[ \frac{1}{n}\sum_{i=1}^{n}\widehat{R}_{i, h-1} \times \boldsymbol{1}\{f (\boldsymbol{X}_{i}) = l\} \right]^2}{\frac{1}{n}\sum_{i=1}^{n}\boldsymbol{1}\{f (\boldsymbol{X}_{i}) = l\}}
    \end{split}
\end{equation}
is bounded away from zero, given the training sample $\{\boldsymbol{X}_i, Y_i\}_{i=1}^{n}$, feature dimension $p$, and network architecture including neuron counts $p_{l}$'s and depth $L$. The second equality follows from the 2-sparse activation assumption and $p_{l} \ge 2^{L-l} M $.

To illustrate the role of Lemma~\ref{lemma10} in deriving this result, consider its application to show that sampling an optimal weight vector and bias for $f_{1,j}$, for any $j \in \{1, \dots, p_1\}$, occurs with positive probability.
Suppose there exists $(\vv{w}_{1,j}^{\star}, c_{1,j}^{\star})$ in space~\eqref{restriction.2}, and we aim to sample $(\vv{w}, c)$ from the same space such that
\[
\boldsymbol{1}\{\vv{w}^{\top} \boldsymbol{X}_i > c\} =
\boldsymbol{1}\{\vv{w}_{1,j}^{\star\top} \boldsymbol{X}_i > c_{1,j}^{\star}\},
\quad \text{ for each } i \in \{1, \dots, n\}.
\]
By Lemma~\ref{lemma10} with $\mathcal{N} = \{\boldsymbol{X}_i\}_{i=1}^{n}$, this event occurs with positive probability. 
The same reasoning extends to each $f_{l,h}$ for $l> 1$, with some $\mathcal{N} \subseteq \{0,1\}^{p_{l-1}}$. 
Finally, since the right-hand side of~\eqref{prob.1} involves at most $\sum_{l=1}^{L} p_{l}$ pairs of weights and biases, multiplying these positive probabilities yields a strictly positive result, establishing the results as in \eqref{prob.1}.

On the other hand, let $T_b$ be the set of idle neurons at the end of the $b$th update round (recall that idle neurons are those not connected to any output neuron). Let $A_b = \{f_{L-1, a_1}, f_{L-1, a_2}\}$ denote two randomly sampled neurons from $T_b$. The existence of at least two neurons at the $(L-1)$th layer follows from the assumption $p_{L-1} \ge 2(M+1)$ and the sparse network structure. Define the event $Q_b$ as follows: at the $(b+1)$th update of $f_{L,h}$, the pair $(\vv{w}, c)$ is in the candidate set (recall there are $K$ randomly sampled weight vectors and biases from \eqref{restriction.2}), where $\vv{w} \in \mathbb{R}^{p_{L-1}}$ satisfies $\|\vv{w}\|_2 = 1$ and $\|\vv{w}\|_0 \le 2$, and $c \in \{\vv{w}^{\top} \vv{e} : \vv{e} \in \{0,1\}^{p_{L-1}}\}$. 
Moreover, $\vv{w}$ assigns nonzero weights only to $A_b$, and  
\[
\boldsymbol{1}\{\vv{w}^{\top} \vv{f}_{L-1}(\vv{x}) > c\} = g(\vv{x}),
\]
where $g$ is from \eqref{prob.1}, and that $\vv{f}_{L-1}(\vv{x})  = (f_{L-1, 1}(\vv{x}), \dots, f_{L-1, p_{L-1}}(\vv{x}) )^{\top}$.

By the definition of the sample-level optimization procedure in Section~\ref{Sec2.2.1} and the results of \eqref{prob.2}, on the event $Q_{b_1 + r}$ for any $r\ge 0$, it holds that $f_{L, h}$ is optimized in the sense of \eqref{eq.2} \textit{after} the $(b_1 + r + 1)$th round. In addition, by similar arguments to \eqref{prob.1}, that
\begin{equation}
    \label{explore.1}
    \begin{split}        
    & \text{there are always sufficient idle neurons in $\widehat{\mathcal{G}}_h$ to construct subnetworks} \\
    & \text{ matching the size of those in $\widehat{\mathcal{G}}$, }
    \end{split}
\end{equation}
 and noting (i) that we randomly sample $K \ge 1$ weights and biases from \eqref{restriction.2} for updating each output neuron, and (ii) that we randomly refresh idle neurons, we conclude that the probability of $Q_{b_1 + r}$ occurring for each $r \ge 0$ is (uniformly) bounded away from zero, depending on the training sample $\{\boldsymbol{X}_i, Y_i\}_{i=1}^{n}$, feature dimension $p$, and network architecture. The proof of~\eqref{explore.1} is deferred to the end of the proof of Theorem~\ref{lemma2}. Moreover, since the events $Q_{b_1}, Q_{b_1+1}, \dots$ are independent, the probability of the complement of $\bigcup_{r=0}^R Q_{b_1 + r}$ tends to zero as $R \to \infty$.

Combining the above arguments, we conclude that $f_{L,h}$ is optimized in the sense of \eqref{eq.2} with high probability as $b$ increases. With $p_L = M$ fixed, a recursive application of the analysis completes the proof of Theorem~\ref{lemma2}.

\begin{proof}[Proof of \eqref{explore.1}] Since there are $M$ output neurons at the $L$th layer and each connects to at most $2^{L-l}$ neurons at layer $l$, the number of active neurons in layer $l$ is at most $2^{L-l}M$.
By this result and the assumption $p_l \ge 2^{L-l}(M+1)$, the number of idle neurons at layer $l$ is bounded below by
\[
p_l - 2^{L-l}M \ge 2^{L-l},
\]
ensuring that at least $2^{L-l}$ idle neurons are always available for subnetwork exploration, as stated in~\eqref{explore.1}.
\end{proof}
        
\subsection{Proof of Example~\ref{example1}}\label{proof.example1}

First, define the event $E_n$ such that, on $E_n$, with probability approaching one as $b \to \infty$, \eqref{eq.2} holds such that for each $j \in \{1, \dots, R_0\}$,
\begin{equation}
    \label{exmp.4}
    f_{L,j}(\boldsymbol{X}) = \boldsymbol{1}\{ X_{j} > 0 \} \quad \text{almost surely},
\end{equation}
where $f_{L,1}, \dots, f_{L,p_{L}}$ are the output neurons of the sample o1Neuro. By Theorem~\ref{lemma2}, with probability approaching one as $b \to \infty$, the sample o1Neuro model optimization achieves the desired solution in the sense of \eqref{eq.2}. Thus, \eqref{exmp.4} specifies a property that must hold on $E_n$ for the sample optimal o1Neuro model. Note that \eqref{exmp.4} concerns the feature vector $\boldsymbol{X}$ at the population level. To show that $\mathbb{P}(E_n^c) \to 0$ as $n \to \infty$, we argue as follows.  
(i) A sample o1Neuro model satisfying \eqref{restriction.2} must also satisfy \eqref{exmp.4} to minimize the \emph{population} loss.  
(ii) When the sample moments in \eqref{eq.2} accurately approximate their population counterparts in \eqref{eq.1}, such a model also satisfies \eqref{exmp.4} to minimize the \emph{sample} loss.  
(iii) The sample moments in \eqref{eq.2} converge to their population counterparts in \eqref{eq.1} with high probability as $n$ increases. Combining (i)--(iii), we conclude that $\mathbb{P}(E_n^c) \to 0$ as $n \to \infty$.

In what follows, we prove (i) above, and begin with $j=1$. 
Under the distributional assumptions of Example~\ref{example1}, it follows that  
\begin{equation}
    \label{exmp.5}
    \mathbb{P}\big(f_{L,1}(\boldsymbol{X}) = 1 \mid \boldsymbol{X}_{-1}\big) \in \{0, 1, 0.5\}
\end{equation}
with probability one for every $f_{L,1}$ satisfying \eqref{restriction.2}.  
If not, suppose there exist $\vv{x}_{-1} \in \{0, 1\}^{p-1}$ and $a \notin \{0, 1, 0.5\}$ such that  $\mathbb{P}\big(f_{L,1}(\boldsymbol{X}) = 1 \mid \boldsymbol{X}_{-1} = \vv{x}_{-1}\big) = a$.
Then
\begin{equation*}
    \begin{split}
        & a \times 0.5^{p-1} \\
        & = \mathbb{P}\big(f_{L,1}(\boldsymbol{X}) = 1 \mid \boldsymbol{X}_{-1} = \vv{x}_{-1}\big) \times \mathbb{P}\big(\boldsymbol{X}_{-1} = \vv{x}_{-1}\big) \\
& = \mathbb{P}\big(\{f_{L,1}(\boldsymbol{X}) = 1\} \cap \{\boldsymbol{X}_{-1} = \vv{x}_{-1}\}\big)  \\
& = \mathbb{P}\big(\{X_j \in E\} \cap \{\boldsymbol{X}_{-1} = \vv{x}_{-1}\}\big)
    \end{split}
\end{equation*}
for some $E \subset \{0,1\}$, contradicting the distributional assumptions where $\mathbb{P}\big(\{X_j \in E\} \cap \{\boldsymbol{X}_{-1} = \vv{x}_{-1}\}\big) = \mathbb{P}(X_j \in E ) \times 0.5^{p-1} \not = a \times 0.5^{p-1} $.

Now, using \eqref{exmp.5}, we show that the function $f_{L,1}$ minimizing the population loss must satisfy 
\begin{equation}
    \label{exmp.32}
    \mathbb{P}\!\left[\mathbb{P}\big(f_{L,1}(\boldsymbol{X}) = 1 \,\big|\, \boldsymbol{X}_{-1}\big) \in \{0,1\}\right] = 0.
\end{equation}
 First, if $\mathbb{P}[\mathbb{P}\big(f_{L,1}(\boldsymbol{X}) = 1 \mid \boldsymbol{X}_{-1}\big) \in \{0, 1\}] >0$, then by our distributional assumptions, 
 \begin{equation}
     \label{exmp.31}
     \mathbb{P}[\mathbb{P}\big(f_{L,1}(\boldsymbol{X}) = 1 \mid \boldsymbol{X}_{-1}\big) \in \{0, 1\}] \ge \min_{\vv{x}_{-1}\in \{0, 1\}^{p-1}}\mathbb{P}(\boldsymbol{X}_{-1} = \vv{x}_{-1} ).
 \end{equation}
 Therefore, it holds that the population $L_2$ loss results from regressing $\beta_1 X_1$ on $H(a_0, a_1) \coloneqq \sum_{l=0}^1 a_l \times \boldsymbol{1}\{f_{L, 1}(\boldsymbol{X}) = l\}$ is lower bounded by
\begin{equation}
    \label{exmp.2}
    \begin{split}        
    & \inf_{ (a_{0}, a_{1} ) \in\mathbb{R}^2} \textnormal{Var}( \beta_1 X_1 - H(a_0, a_1) ) \\
     & \ge \mathbb{E}[\inf_{ (a_{0}, a_{1} ) \in\mathbb{R}^2}\textnormal{Var}( \beta_1 X_1 - H(a_0, a_1) \mid \boldsymbol{X}_{-1})] \\
     & \ge \mathbb{E}[ \textnormal{Var}( \beta_1 X_1  \mid \boldsymbol{X}_{-1}) \times \boldsymbol{1}\{\mathbb{P}\big(f_{L,1}(\boldsymbol{X}) = 1 \mid \boldsymbol{X}_{-1}\big) \in \{0, 1\}\}] \\
     & \quad \quad + \mathbb{E}[ 0\times \boldsymbol{1}\{\mathbb{P}\big(f_{L,1}(\boldsymbol{X}) = 1 \mid \boldsymbol{X}_{-1}\big) =0.5\}] \\
     & \ge \frac{\beta_{1}^2}{4} \times \mathbb{P}(\mathbb{P}\big(f_{L,1}(\boldsymbol{X}) = 1 \mid \boldsymbol{X}_{-1}\big) \in \{0, 1\} ) \\
     & \ge \frac{\beta_{1}^2}{4} \times \min_{\vv{x}_{-1}\in \{0, 1\}^{p-1}}\mathbb{P}(\boldsymbol{X}_{-1} = \vv{x}_{-1} ) \\
     & \ge 2^{-p+1} \times \frac{\beta_{1}^2}{4},
     \end{split}
\end{equation}
where the fourth inequality follows from \eqref{exmp.31}.
On the other hand, if \eqref{exmp.32}, which implies that $f_{L, 1}(\boldsymbol{X}) = \boldsymbol{1}\{X_{1} > 0\}$ almost surely due to our distributional assumptions and \eqref{exmp.5}, the population loss is upper bounded by
\begin{equation}
    \label{exmp.3}
   \sum_{l=2}^{R_0} \textnormal{Var}( \beta_l X_l ) = \frac{\sum_{l=2}^{R_0}\beta_{l}^2}{4}.
\end{equation}

By our assumption $\gamma = 1$, \eqref{exmp.2}--\eqref{exmp.3}, and similar arguments as for \eqref{exmp.3}, and
$$\zeta_j \coloneqq 2^{-p+1} \beta_j^2 - \sum_{l=j+1}^{R_0} \beta_{l}^2> 0$$ 
due to the assumption $2^{-p+1} \beta_j^2 > \sum_{l=j+1}^{R_0} \beta_{l}^2$ for each $j\in \{1, \dots, R_0\}$, we deduce that the first optimal neuron satisfies
\[
f_{L,1}(\boldsymbol{X}) = \boldsymbol{1}\{ X_{1} > 0 \} \quad \text{almost surely}.
\]

By recursively applying the above arguments for each $j\in \{1, \dots, R_0\}$, we conclude that, to minimize the population loss, the sample optimal o1Neuro satisfies \eqref{exmp.4}.

Now, let us turn to (ii) and (iii) that deals with statistical estimation.  With sufficiently many samples, applications with standard concentration inequalities and the assumptions that $R_0$ is finite with $\min_{1\le j\le R_0 }\zeta_{j}>0$ show that with probability approaching one as $n\rightarrow \infty$, it holds that $\widetilde{a}_{0h}(f)$, $\widetilde{a}_{1h}(f)$ and $\widetilde{W}_{h}(f)$ for each $f\in\widetilde{\mathcal{G}}$ and  every $h\in \{1, \dots, R_0\}$  can be accurately estimated by $\widehat{a}_{0h}(f)$, $\widehat{a}_{1h}(f)$ and $\widehat{W}_{h}(f)$ with high precision such that the sample optimal output neurons follows \eqref{exmp.4} as well. Therefore,
\begin{equation}
    \label{exmp.6}
    E_n \text{ occurs with arbitrarily high probability } \varepsilon_n \le 1, \text{ for sufficiently large } n,
\end{equation}
which concludes \eqref{exmp.4}. The detailed applications of standard concentration inequalities are omitted for brevity.


In what follows, we calculate the probability that $f_{L,j}(\boldsymbol{X}) = \boldsymbol{1}\{X_{j} > 0\}$ after a given number of update rounds, assuming that $\boldsymbol{1}\{X_{j} > 0\}$ is the optimal solution for the $j$th output neuron at both the sample and population levels. For some small $\varepsilon_0 > 0$, define
$$W_{n} = \{\min_{1\le j\le R_0}\texttt{\#}\{X_{ij} = 0\} \ge  n(\frac{1}{2} - \varepsilon_0)\}.$$

First, when $f_{1,h}$ is an idle neuron at the end of the $b_1$th round for some integer $b_1$, the probability that $f_{1,h}(\boldsymbol{X}) = \boldsymbol{1}\{X_{j} > 0\}$ is at least the probability of simultaneously having (1) $\lVert \vv{w}_{1,h} \rVert_0 = 1$, (2) the $j$th coordinate of $\vv{w}_{1,h}$ equal to one, and (3) $c_{1,h} = 0$. This yields a probability of $\frac{1}{2} \times \frac{1}{p} \times (\frac{1}{2} - \varepsilon_0) \ge \frac{1}{2p} (\frac{1}{2} -  \varepsilon_0)$, conditional on $W_n$, by our parameter sampling scheme in~\eqref{restriction.2}.  Define the event $B_1$ as the occurrence that at least 
\[
p_1 \times \Big( \frac{1}{2p} \Big( \frac{1}{2} - \varepsilon_0 \Big) - \varepsilon_1 \Big)
\] 
first-layer neurons are equivalent to $\boldsymbol{1}\{X_j > 0\}$ at the end of the $b_1$th round, for some small $\varepsilon_1 > 0$. Since $L$ and $p_L$, which set the upper limits on the number of idle neurons per layer, are fixed, the probability of $B_1$ is at least $\delta_1$, which can be made arbitrarily close to one by taking $p_1$ sufficiently large.

Next, consider the second layer, conditional on $W_n \cap B_1$. When $f_{2,h}$ is an idle neuron at the end of the $b_1$th round, the probability that $f_{2,h}(\boldsymbol{X}) = \boldsymbol{1}\{X_{j} > 0\}$ is at least the probability of simultaneously having 
(1) $\lVert \vv{w}_{2,h} \rVert_0 = 1$, 
(2) $f_{1,j}(\boldsymbol{X}) = \boldsymbol{1}\{X_{j} > 0\}$ and that the $j$th coordinate of $\vv{w}_{2,h}$ is one, and 
(3) $c_{2,h} = 0$, 
which gives 
\begin{equation}
    \label{exmp.8}
    \frac{1}{2} \times \Big[\frac{1}{p_1}\times p_1\times \Big(\frac{1}{2p} \Big(\frac{1}{2} - \varepsilon_0 \Big) - \varepsilon_1 \Big)\Big] \times \frac{1}{2} =  4^{-2} p^{-1} - \frac{\varepsilon_0}{8p} - \frac{\varepsilon_1}{4},
\end{equation}
according to our parameter sampling scheme in~\eqref{restriction.2}. Consequently, we define the event $B_2$ where at least ($p_2$ times the left-hand side of \eqref{exmp.8})
\[
p_2 \times \Big[ \frac{1}{4} \Big(\frac{1}{2p} \Big(\frac{1}{2} - \varepsilon_0 \Big) - \varepsilon_1 \Big) - \varepsilon_2 \Big]
\] 
neurons in the second layer are equivalent to $\boldsymbol{1}\{X_{j} > 0\}$, for some small $\varepsilon_2 > 0$. For sufficiently large $p_2$, the probability of $B_2$ is $\delta_2$ conditional on $W_n\cap B_1$, which can be made arbitrarily close to one.

Applying the arguments recursively up to the $L$th layer, for sufficiently large $\min_{1 \le l < L} p_l$, the probability conditional on $W_n \cap B_1 \cap\dots \cap B_{L-1}$ that some randomly sampled candidate $(\vv{w}, c)$ satisfies $\boldsymbol{1}\{\vv{w}^{\top}\vv{f}_{L-1}(\boldsymbol{X} )> c\} = \boldsymbol{1}\{X_j > 0\}$ is at least
\begin{equation}
    \label{exmp.7}
    \left( 4^{-L} p^{-1} - \varepsilon_L  \right),
\end{equation}
for some small $\varepsilon_L > 0$ depending on $\varepsilon_0, \dots, \varepsilon_{L-1}$, conditional on $W_n $. The derivation of \eqref{exmp.7} follows a similar reasoning as the simplified example shown in \eqref{exmp.8}. When this event and $E_n$ both occur, the sample-level optimization procedure in Section~\ref{Sec2.2.1} ensures that $f_{L,j}(\boldsymbol{X}) = \boldsymbol{1}\{X_j > 0\}$ at the end of the $(b_1+1)$th round and \textit{remains so in all subsequent rounds}.

By \eqref{exmp.7} and using similar arguments to those employed in its derivation, the probability that 
$f_{L,j}(\boldsymbol{X}) \neq \boldsymbol{1}\{X_j > 0\}$ 
during the rounds from $b_1$ to $(b_1 + 4^{L} p \kappa)$ is bounded above by  
\[
\left( 1 - K\Big(4^{-L} p^{-1} - \varepsilon_{L} - \sum_{l=1}^{L-1}\big(1 - \delta_{l}\big) \Big) \right)^{4^{L} p \kappa},
\]
conditional on $W_n \cap E_n$, for any $b_1 > 0$.  
Here, the subtraction of $\sum_{l=1}^{L-1}(1 - \delta_l)$ accounts for the fact that the events $B_1, \dots, B_{L-1}$ are not conditioned upon, and  $K$ is the number of randomly sampled candidates.

Lastly, after $4^{L} \kappa R_0 p$ rounds of updates, the probability of completing the optimization in the sense of \eqref{eq.2} for all $j \in \{1, \dots, R_0\}$ is at least
\[
1 \;-\; R_0 \left( 1 - K \Big( 4^{-L} p^{-1} - \varepsilon_{L} - \sum_{l=1}^{L-1} (1 - \delta_{l}) \Big) \right)^{4^{L} p \kappa} 
\;-\; \mathbb{P}(E_n^c) 
\;-\; \mathbb{P}(W_n^c).
\]

Using  the exponential inequality $1-x < e^{-x}$ for $x>0$, the probability is further lower bounded by
\[
1 - R_0 e^{-\kappa K}
\]
for sufficiently small $\varepsilon_{L}, \mathbb{P}(E_n^c)\le 1 - \varepsilon_n, \sum_{l=1}^{L-1}(1 - \delta_{l})$, and $\mathbb{P}(W_n^c)$ with fixed $(R_0, K , \kappa, p, L)$.  Note that $\varepsilon_{L}$, $1 - \varepsilon_n$, $\sum_{l=1}^{L-1}(1 - \delta_{l})$, and $\mathbb{P}(W_n^c)$ can all be made arbitrarily small by first choosing sufficiently large $\min_{1 \le l < L} p_{l}$ and then taking $n$ sufficiently large.

We thus conclude the proof of Example~\ref{example1}.

\renewcommand{\theequation}{C.\arabic{equation}}
	\renewcommand{\thesubsection}{C.\arabic{subsection}}
	\setcounter{equation}{0}

\section{Proofs of Technical Lemmas}\label{SecC}
\subsection{Proof of Lemma~\ref{lemma3}}\label{proof.lemma3}
    
We begin the proof by demonstrating that the class $\mathcal{D}$ includes all linear combinations of finitely many rectangular indicator functions, where each indicator function depends on only $k$ coordinates. Note that it suffices to show that an element of $\widetilde{\mathcal{G}}$  with  $L= 2 + \lceil \log_{2} k \rceil$ can approximate any rectangular indicator function depending on $k$-coordinates. Specifically, for any rectangle $R' = [a_1, b_1] \times \cdots \times [a_k, b_k] \subset [0, 1]^k$, any index set $\{i_1, \dots, i_k\} \subset \{1, \dots, p\}$, and the corresponding set
$R = \left\{ \vv{x} \in [0,1]^p : (x_{i_1}, \dots, x_{i_k}) \in R' \right\},$ we construct element of $\widetilde{\mathcal{G}}$ with  $L= 2 + \lceil \log_{2} k \rceil$ as follows. The following construction adheres to 
network parameter space \eqref{restriction.1}.

    For the first and second layers, we construct that $f_{2, q}(\vv{x}) = \boldsymbol{1}\{\frac{\sqrt{2}}{2}f_{1, 2q - 1} (\vv{x})- \frac{\sqrt{2}}{2}f_{1, 2q} (\vv{x}) >0\}$,  $f_{1, 2q-1}(\vv{x}) =\boldsymbol{1}\{ x_{i_{q}}> a_q\}$, and $f_{1, 2q}(\vv{x}) = \boldsymbol{1}\{ x_{i_{q}}> b_q\}$ for $q\in \{1, \dots, k\}$. Eventually, we shall see that there are exactly $a_2 = k$ and $a_1 = 2k$ active neurons respectively at the second and first layers.

    For the third layer, we construct that $f_{3, r}(\vv{x}) = \boldsymbol{1}\{\frac{\sqrt{2}}{2} f_{2, 2r - 1} (\vv{x}) +  \frac{\sqrt{2}}{2} f_{2, 2r} (\vv{x}) > \frac{\sqrt{2}}{2} \}$ for $r \in \{1, \dots, \lfloor\frac{k}{2} \rfloor\}$ and $f_{3, \lceil\frac{k}{2} \rceil }(\vv{x}) = \boldsymbol{1}\{f_{2, k} (\vv{x})  > 0 \}$ if $\frac{k}{2} > \lfloor\frac{k}{2} \rfloor$. Eventually, we shall see that there are at most $a_3 = \lceil\frac{k}{2} \rceil = \lceil\frac{a_2}{2} \rceil$ active neurons at the third layer. The $l$th layer, for $l > 3$, is defined in a manner analogous to the third layer, implying that there are at most $a_l = \lceil \frac{a_{l-1}}{2} \rceil$ active neurons at layer $l$.

We repeat this construction until we reach the output layer and construct $f_{L, 1}(\vv{x}) = \boldsymbol{1}\{\frac{\sqrt{2}}{2} f_{L-1, 1}(\vv{x}) + \frac{\sqrt{2}}{2} f_{L-1, 2}(\vv{x}) > \frac{\sqrt{2}}{2}\}$, where we define $a_L = 1$. A simple calculation shows that $L = 2 + \lceil \log_2 k \rceil = \min\{ l : a_l = 1 \}$, giving the minimum depth required for the constructed network $f_{L,1}(\vv{x})$.

Now, the construction of $f_{L, 1}(\vv{x})$ forms the desired rectangular indicator function, yielding
$$\int_{[0, 1]^p} \Big| f_{L, 1}(\vv{x}) - \boldsymbol{1}\{\vv{x} \in R\} \Big| d\vv{x} = 0,$$
and implying that
$$\inf_{f \in \widetilde{\mathcal{G}}} \int_{[0, 1]^p} \Big| f(\vv{x}) - \boldsymbol{1}\{\vv{x} \in R\} \Big| d\vv{x} = 0.$$
Such a results and the definition of $\mathcal{D}$ conclude that $\mathcal{D}$ includes all linear combinations of finitely many rectangular indicator functions, where each indicator function depends on only $k$ coordinates.

On the other hand, it is well known that $p$-dimensional step functions, defined as linear combinations of finitely many rectangular indicator functions, each depending on at most $p$ coordinates, can approximate any measurable function on a compact domain, such as $[0, 1]^p$, in the Lebesgue measure sense~\citep{stein2009real, cohn2013measure}. However, since our focus is on the distribution of $\boldsymbol{X}$ rather than the Lebesgue measure, we assume in Condition~\ref{condi.inde} that $\boldsymbol{X}$ has a bounded probability density function to apply existing results. Under this assumption, there exists a constants $C_2 > 0$ such that for any measurable function $f: [0, 1]^p \to \mathbb{R}$,
$$
 \mathbb{E}[f^2(\boldsymbol{X})] \le C_2 \int_{[0,1]^p} f^2(\vv{x})\, d\vv{x}.
$$
This equivalence between $L_2$-norms under the probability measure and the Lebesgue measure allows us to apply standard approximation results~\citep{cohn2013measure, stein2009real}.

In particular, by arguments similar to those in Proposition 3.4.3 of~\citep{cohn2013measure} and Theorem 4.3 of~\citep{stein2009real}, the class $\overline{\mathcal{D}}$ contains $\mathcal{L}(k)$. That is, for every $f \in \mathcal{L}(k)$, there exists a sequence $f_i \in \mathcal{D}$ such that $\lim_{i \to \infty} \mathbb{E}\big[(f(\boldsymbol{X}) - f_i(\boldsymbol{X}))^2\big] = 0$. Here, we highlight two extensions in our setting in comparison to the similar results presented in \citep{stein2009real, cohn2013measure}: first, we consider the distribution of $\boldsymbol{X}$ instead of the Lebesgue measure, and second, we account for $k$-sparse measurable functions in $\mathcal{L}(k)$. Nevertheless, with Condition~\ref{condi.inde}, the standard arguments in the cited references apply directly, and the detailed proof is omitted for simplicity.

To finish the proof of Lemma~\ref{lemma3}, note that by the previous results it holds that for each $f_r\in\mathcal{L}(k)$, there exists $f_{i}^{(r)}\in \mathcal{D}$ such that $\mathbb{E}\big[(f_r(\boldsymbol{X}) - f_i^{(r)}(\boldsymbol{X}))^2\big]$ approaches zero. By this result and an application of the Minkowski inequality, it holds that 
$$\sqrt{ \mathbb{E} \{ [ \sum_{r=1}^{R_0} f_r(\boldsymbol{X}) ] - \sum_{r=1}^{R_0} f_i^{(r)}(\boldsymbol{X})\}^2}\le \sum_{r=1}^{R_0} \sqrt{\mathbb{E}\big[(f_r(\boldsymbol{X}) - f_i^{(r)}(\boldsymbol{X}))^2\big]} \rightarrow 0 \textnormal{ as } i \rightarrow \infty,$$
which, in combination with the facts that $\sum_{r=1}^{R_0} f_r \in \mathcal{L}(k, R_{0})$ and $\sum_{r=1}^{R_0} f_i^{(r)} \in \mathcal{D}$, concludes the desired results that $\overline{\mathcal{D}}$ contains $\mathcal{L}(k, R_{0})$. We have  completed the proof of Lemma~\ref{lemma3}.

\subsection{Lemma~\ref{lemma10} and its Proof}\label{proof.lemma10}

Lemma~\ref{lemma10} shows that when randomly sampling a weight vector from 
$\{\vv{w} \in \mathbb{R}^k : \|\vv{w}\|_2 = 1, \|\vv{w}\|_0 \le w_0\}$, there exists a positive probability of obtaining one that is equivalent, in the sense of~\eqref{sample-euquivalent}, to a given target weight vector $\vv{w}$ from the same space. Lemma~\ref{lemma10} follows from the zero-gradient property of indicator activations.

\begin{lemma}\label{lemma10}
		
		Let $w_0$ be a constant integer with $1 \le w_0 \le k$, and let $\mathcal{N}$ be a set of input vectors in $[0,1]^k$. There exists some constant $\varepsilon>0$ such that for every pair $(\vv{w}, c)\in \{(\vv{u}, b) : \vv{u} \in\mathbb{R}^k, \norm{\vv{u}}_2= 1, \norm{\vv{u}}_0\le w_0, b\in\mathbb{R}\}$ with $\texttt{\#} \{\vv{u} \in \mathcal{N}: \vv{w}^{\top}\vv{u} - c > 0\} < \texttt{\#}\mathcal{N}$, there is a measurable set of  weight vectors $\mathcal{E}(\vv{w}, c)\subset\{\vv{w} \in\mathbb{R}^k: \norm{\vv{w}}_2= 1, \norm{\vv{w}}_0\le w_0\}$ with at least positive surface measure $\varepsilon$ on $\{\vv{w} \in\mathbb{R}^k: \norm{\vv{w}}_2= 1, \norm{\vv{w}}_0\le w_0\}$ such that for every $\vv{w}^{'} \in \mathcal{E}(\vv{w}, c)$, it holds that 
        \begin{equation}
            \label{sample-euquivalent}
            \{\vv{u} \in \mathcal{N}: (\vv{w}^{'})^{\top}\vv{u} - (\vv{w}^{'})^{\top}\vv{v} > 0\} = \{\vv{u} \in \mathcal{N}: \vv{w}^{\top}\vv{u} - c > 0\}
        \end{equation} 
        for some $\vv{v}\in \mathcal{N}$.

	\end{lemma}

\noindent\textit{Proof of Lemma~\ref{lemma10}: } \quad     We first consider the scenario where $w_0 = k$, and begin the proof by showing that, for a given pair $(\vv{w}, c)$, there exists a set of weight–bias pairs $\mathcal{A}(\vv{w}, c)$ such that \textnormal{(1)} $\{ \vv{w}' : (\vv{w}', c') \in \mathcal{A}(\vv{w}, c) \}$ is measurable with positive surface measure  on the  $k$-dimensional unit sphere, and \textnormal{(2)} $\{\vv{u} \in \mathcal{N}: (\vv{w}^{'})^{\top}\vv{u} - c^{'} > 0\} = \{\vv{u} \in \mathcal{N}: \vv{w}^{\top}\vv{u} - c > 0\}$ for every $(\vv{w}^{'}, c^{'}) \in \mathcal{A}(\vv{w}, c)$.
		
		In what follows, we prove the above claim. First, 
		define $\delta>0$ such that $2\delta = \min\{1, \min \{ a  :\vv{u}\in\mathcal{N}, a = \vv{w}^{\top} \vv{u} - c > 0 \}\}$, where we define $\min\{ \emptyset\} = \infty$. Then, $\{ \vv{u}\in\mathcal{N}: \vv{w}^{\top} \vv{u} - c - \delta >0\} = \{ \vv{u}\in\mathcal{N}: \vv{w}^{\top} \vv{u} - c > 0\} \eqqcolon\mathcal{N}_{+}$. Additionally, define 
		$$\mathcal{A}(\vv{w}, c) = \{ (\vv{w}^{'}, c^{'}): \norm{\vv{w}^{'} - \vv{w}}_2 \le \frac{\delta}{4\sqrt{k}}, \norm{\vv{w}^{'} }_2 = 1 , |c^{'} - (c + \frac{1}{2}\delta)| < \frac{\delta}{4}\},$$
		and notice that $\mathcal{A}(\vv{w}, c)$ satisfies the first requirement listed above: $\{ \vv{w}' : (\vv{w}', c') \in \mathcal{A}(\vv{w}, c) \}$ is measurable  with some positive surface measure $\varepsilon(\vv{w}, c) > 0$ on the  $k$-dimensional unit sphere.
		
		For each $\vv{u}\in \mathcal{N}_{+}$ and $(\vv{w}^{'}, c^{'}) \in \mathcal{A}(\vv{w}, c)$, it holds that 
		$$(\vv{w}^{'})^{\top} \vv{u} - c^{'} > \vv{w}^{\top} \vv{u}  - \frac{\delta}{4} - (c+ \frac{\delta}{2}) - \frac{\delta}{4}\ge \min\{\vv{w}^{\top} \vv{u} - c - \delta, \vv{w}^{\top} \vv{u} - c \} > 0, $$
		where the first inequality holds because  $\norm{ (\vv{w}^{'} - \vv{w})^{\top} \vv{u} }_2 \le \frac{\delta}{4\sqrt{k}} \sqrt{k}$ due to the Cauchy-Schwarz inequality (recall that $\mathcal{N}$ consists of elements from $[0, 1]^k$, and therefore $\norm{\vv{u}}_2 \le \sqrt{k}$) and $|c^{'} - (c + \frac{1}{2}\delta)| < \frac{\delta}{4}$. The second and third inequalities result from the definition of $\delta$. By these results and the definition of $\mathcal{N}_{+}$, it holds that  $\mathcal{A}(\vv{w}, c)$ satisfies the second requirement listed above: $\{\vv{u} \in \mathcal{N}: (\vv{w}^{'})^{\top}\vv{u} - c^{'} > 0\} = \{\vv{u} \in \mathcal{N}: \vv{w}^{\top}\vv{u} - c > 0\}$ for every $(\vv{w}^{'}, c^{'}) \in \mathcal{A}(\vv{w}, c)$.
		
		Next, note that the above arguments applies to each distinct sample separation, and that there are at most $2^{\texttt{\#}\mathcal{N}}$  distinct ways to separate $\mathcal{N}$ into two subsamples by hyperplanes. For each separation, we can pick up a representative split. Eventually, we collect a finite set of representative splits, denoted by $\{(\vv{w}_{l}^{\dagger}, c_{l}^{\dagger})\}_{l=1}^{L_0}$ for some integer $L_0>0$. Furthermore, each of these splits correspond to a measure lower bound $\varepsilon(\vv{w}_{l}^{\dagger}, c_{l}^{\dagger}) >0$. As there are finitely many distinct separations, we conclude that $\varepsilon = \min_{1\le l\le  L_0} \varepsilon(\vv{w}_{l}^{\dagger}, c_{l}^{\dagger}) >0$.

		Define $\mathcal{E}(\vv{w}, c) = \{ \vv{w}' : (\vv{w}', c') \in \mathcal{A}(\vv{w}, c) \}$. To conclude the proof, let us observe that for each pair $(\vv{w}, c)$, there exists some $\vv{v}\in \mathcal{N}$ such that $\{\vv{u} \in \mathcal{N}: \vv{w}^{\top}\vv{u} - \vv{w}^{\top}\vv{v} > 0\} = \{\vv{u} \in \mathcal{N}: \vv{w}^{\top}\vv{u} - c > 0\}$ when $\texttt{\#} \{\vv{u} \in \mathcal{N}: \vv{w}^{\top}\vv{u} - c > 0\} < \texttt{\#} \mathcal{N}$. By this and the construction of $\mathcal{E}(\vv{w}, c)$, we have completed the proof of Lemma~\ref{lemma10} for the case with $w_0 = k$.

To establish similar results for 
$(\vv{w}, c) \in \{(\vv{u}, b) : \vv{u} \in \mathbb{R}^k, \|\vv{u}\|_2 = 1, \|\vv{u}\|_0 \le w_0, b \in \mathbb{R}\}$ 
with general $1 \le w_0 \le k$, we restrict attention to the nonzero coordinates 
$\{j \mid \vv{w} = (w_1, \dots, w_k)^{\top}, |w_j| > 0\}$ 
and apply the same arguments as above to obtain the same result. The detail is omitted for brevity. 

We have completed the proof of Lemma~\ref{lemma10}.

\end{document}